\documentclass[acmtog]{acmart}
\acmSubmissionID{551}

\usepackage{booktabs} 
\usepackage[ruled]{algorithm2e} 
\usepackage{comment}

\usepackage[capitalise]{cleveref}
\usepackage{multirow}
\usepackage{makecell}
\usepackage[table]{xcolor} 
\usepackage{pifont}

\usepackage{hyperref}

\usepackage{cuted}
\usepackage[table]{xcolor}
\usepackage{setspace}

\SetAlFnt{\small}
\SetAlCapFnt{\small}
\SetAlCapNameFnt{\small}
\SetAlCapHSkip{0pt}

\acmJournal{TOG}
\acmYear{2026}
\acmMonth{1}

\copyrightyear{2026}

\newcommand{\ours}{HL-OutPaint}

\newcommand{\gcg}{GCG}
\newcommand{\gcgfull}{Global Coarse Guidance}

\newcommand{\first}{Global Coarse Guidance Construction}    

\newcommand{\second}{GCG-Guided Video Outpainting}    

\newcommand{\swap}{global-local frame swapping}    

\newcommand{\fix}[1]{\textcolor{black}{#1}}

\newcommand{\link}[1]{\textcolor{blue}{#1}}

\copyrightyear{2026}
\acmYear{2026}
\setcopyright{cc}
\setcctype{by}
\acmConference[SIGGRAPH Conference Papers '26]{Special Interest Group on Computer Graphics and Interactive Techniques Conference Conference Papers}{July 19--23, 2026}{Los Angeles, CA, USA}
\acmBooktitle{Special Interest Group on Computer Graphics and Interactive Techniques Conference Conference Papers (SIGGRAPH Conference Papers '26), July 19--23, 2026, Los Angeles, CA, USA}
\acmDOI{10.1145/3799902.3811098}
\acmISBN{979-8-4007-2554-8/2026/07}

\begin{document}

\title{\fix{{\ours{}: Coarse-to-Fine Video Outpainting for High-Resolution Long-Range Videos}}}


\author{Jeongeun Park}
\affiliation{
  \institution{POSTECH}
  \country{Republic of Korea}
}
\email{koyy001@postech.ac.kr}

\author{Janghyeok Han}
\affiliation{
  \institution{POSTECH}
  \country{Republic of Korea}
}
\email{hjh9902@postech.ac.kr}

\author{Geonung Kim}
\affiliation{
  \institution{POSTECH}
  \country{Republic of Korea}
}
\email{k2woong92@postech.ac.kr}

\author{Hyun-Seung Lee}
\affiliation{
  \institution{Visual Display Business, Samsung Electronics}
  \country{Republic of Korea}
}
\email{hyuns.lee@samsung.com}

\author{Kyuha Choi}
\affiliation{
  \institution{Visual Display Business, Samsung Electronics}
  \country{Republic of Korea}
}
\email{kyuha75.choi@samsung.com}

\author{Youngseok Han}
\affiliation{
  \institution{Visual Display Business, Samsung Electronics}
  \country{Republic of Korea}
}
\email{yseok.han@samsung.com}

\author{Sunghyun Cho}
\affiliation{
  \institution{POSTECH}
  \country{Republic of Korea}
}
\email{s.cho@postech.ac.kr}




\begin{abstract}
Video outpainting generates plausible visual content beyond a video’s original spatial extent, playing a key role in adapting videos to diverse display formats. To support such use cases, it must enable large spatial extrapolation over long sequences. However, most existing methods address only one challenge or lack explicit mechanisms, leaving notable limitations.
\fix{In this paper, we propose \ours{}, a high-resolution video outpainting framework for long sequences.
Our approach follows a coarse-to-fine strategy with a two-stage pipeline.
We first construct a \gcgfull{} (\gcg{}), a low-resolution representation that captures global structure and dominant motion across the video. Unlike na"ive downsampling, the \gcg{} is built via a novel \swap{} mechanism that couples sparse global keyframes with local temporal windows and exchanges information during sampling. This enables the \gcg{} to encode both long-term structural consistency and short-term temporal dynamics in a unified representation.
Guided by this representation, \ours{} then performs high-resolution outpainting to generate spatially detailed and temporally consistent content.
By separating global structure modeling from fine-grained synthesis, our framework achieves stable, coherent generation for large spatial expansion and long video sequences.}
Extensive experiments show that \ours{} outperforms existing methods in challenging scenarios with wide spatial extrapolation and long video sequences. Project page available at \link{\href{https://koyy001.github.io/Publications/hl-outpaint}{https://koyy001.github.io/Publications/hl-outpaint}}.

\end{abstract}



\begin{teaserfigure}
  \centering
  \includegraphics[width=0.882\textwidth]{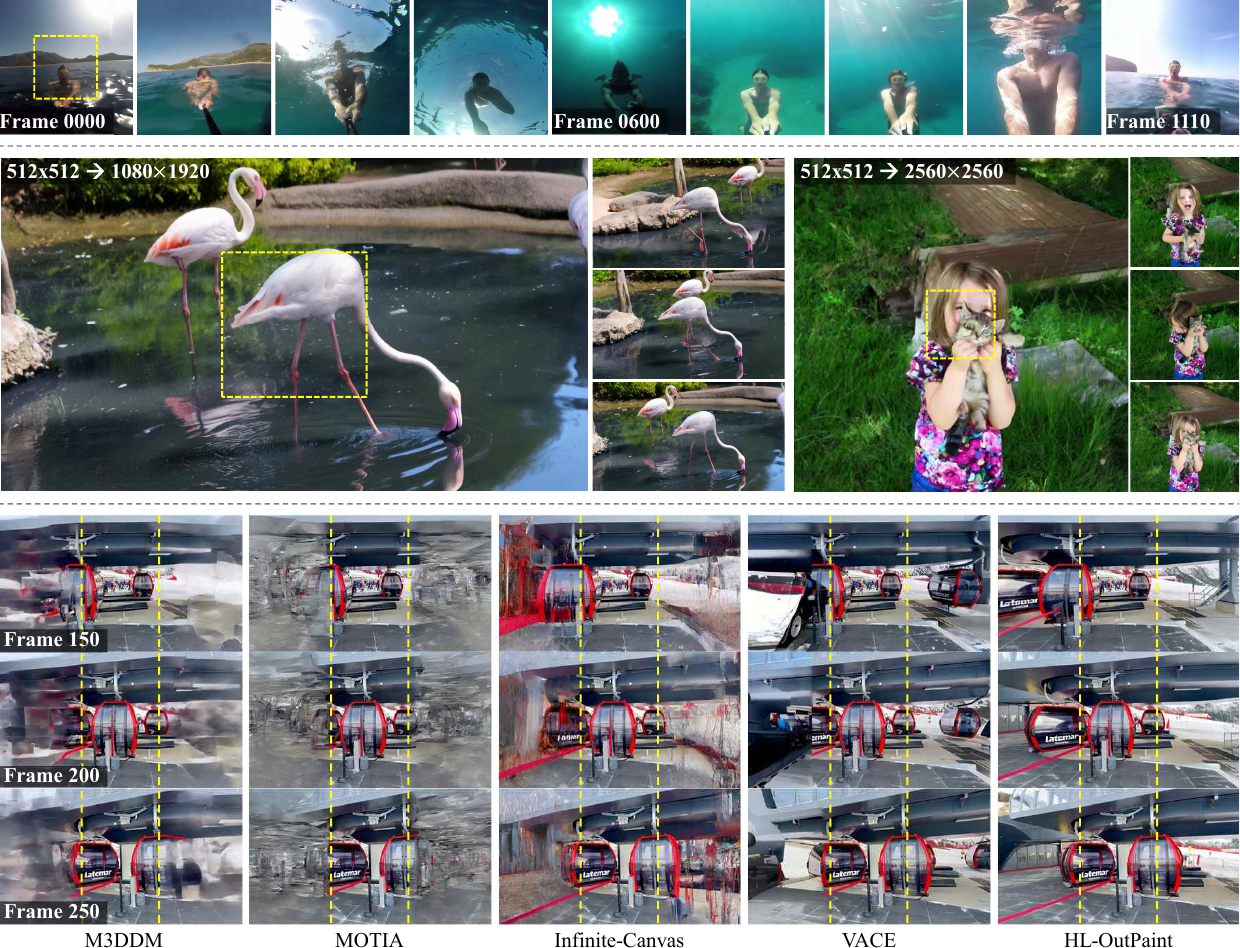}
  \caption{
  \ours{} handles \emph{long‑range} outpainting (top) and \emph{high‑resolution} outpainting (middle), and outperforms existing state‑of‑the‑art methods, including M3DDM~\cite{m3ddm}, MOTIA~\cite{motia}, Infinite‑Canvas~\cite{ic}, and VACE~\cite{vace} (bottom). The yellow dashed boxes indicate the original regions before outpainting. The input videos are from the DAVIS dataset~\cite{davis} (flamingo, cat-girl) and short-Form dataset.
  }
  \Description{teaser}
  \label{fig:teaser}
\end{teaserfigure}

\maketitle

\section{Introduction}



\fix{
Video outpainting aims to synthesize plausible visual content beyond the spatial boundaries of an input video, preserving visual composition even when the original frame provides limited context. 
It is important for adapting fixed-aspect-ratio videos to diverse displays and for editing tasks such as reframing, stabilization, and creating space for overlays. As video consumption spans increasingly diverse devices, flexible and high-quality spatial extension has become essential in modern video production pipelines.
}

To achieve high-quality video outpainting, it is essential to generate temporally and spatially coherent content.
To this end, several approaches have been proposed using generative models, such as generative adversarial networks (GANs) \cite{dehan} and diffusion models \cite{motia, vace, unboxed}, demonstrating impressive outcomes under predefined resolutions and limited video sequence lengths.
However, in practice, video outpainting often requires much larger spatial extrapolation over significantly longer video sequences, revealing a gap between existing approaches and real-world requirements.

Recent methods address only part of this challenge. Infinite-canvas \cite{ic} handles large spatial extrapolation with patch-based tiles, but its local generation strategy limits global coherence and can cause repetitive or inconsistent structures. Conversely, M3DDM \cite{m3ddm} focuses on long video sequences by subsampling frames to form a condensed clip and using the resulting keyframes as long‑term guidance. Yet when the input contains rapid motion, the temporal gap between keyframes becomes large, frequently leading to temporal inconsistencies. Thus, large-scale and long-sequence video outpainting remains challenging, with no existing unified solution addressing both dimensions simultaneously.

\fix{
In this paper, we propose \ours, a unified framework for high-resolution video outpainting over long sequences. To ensure global coherence across large spatio-temporal extents, \ours{} adopts a coarse-to-fine strategy consisting of two stages. The first stage constructs a \gcgfull{} (\gcg), which is a spatio-temporally low-resolution but globally coherent outpainting of the entire video sequence formed by a set of sparse global keyframes.
By constructing \gcg{} in a spatio-temporally reduced resolution, \ours{} allows the diffusion model to optimize the full sequence holistically within its attention span, thereby establishing a consistent structural foundation.
Then, the second stage performs high-resolution refinement using a tile-based diffusion strategy guided by this global structure.
}

\fix{
However, to bring a long-range video into such a single processing pass, an aggressive downsampling is inevitable, which necessitates discarding the majority of intermediate frames. While such a compressed perspective is effective for ensuring global coherence, it inherently loses fine-grained temporal cues, such as objects appearing or disappearing within local windows, that are critical for maintaining motion integrity.
}

\fix{
To bridge this gap, we introduce a novel global-local frame swapping mechanism integrated into the \gcg{} construction process. This mechanism couples sparse global keyframes in the GCG with their corresponding local temporal windows during the diffusion denoising steps. By exchanging information between global keyframes and local temporal windows, global keyframes can inherit detailed temporal observations from local windows that would otherwise be lost in the downsampled representation. This bidirectional flow ensures that the \gcg{} remains both globally stable and locally accurate, providing a robust anchor for high-resolution synthesis.
}

\fix{
Our main contributions are summarized as follows:
\begin{itemize}
    \item We propose \ours{}, a novel video outpainting framework that jointly ensures spatial and temporal coherence for real‑world high‑resolution, long‑range video sequences.
    \item We introduce a \swap{} mechanism that couples sparse keyframes with local temporal windows, enabling both long-range structural consistency and short-range temporal coherence.
    \item Extensive experiments demonstrate that \ours{} achieves state‑of‑the‑art performance across diverse scenarios requiring both wide spatial expansion and long‑range temporal consistency.
\end{itemize}
}

\section{Related Work}

\paragraph{Image/Video Outpainting}
With the advent of generative models, various image outpainting methods have been proposed, including GAN‑based~\cite{InOut,EGP,SEIO,pathakCVPR16context}, diffusion‑based~\cite{Palette,CMIO,VIP,PAO,bridgingthegap}, and masked‑prediction approaches~\cite{maskgit}. Although these methods demonstrate strong spatial extrapolation capabilities, their naive extension to videos causes severe flickering due to the lack of temporal modeling. To apply the techniques to videos, several approaches have been proposed. \cite{dehan} perform background outpainting under the assumption that foreground objects remain within the original frame. MOTIA~\cite{motia} improves coherence through test-time adaptation on the input video, while Unboxed~\cite{unboxed} enforces temporal consistency by reconstructing static regions in 3D. VACE~\cite{vace} leverages large-scale diffusion training to enhance generative quality. \fix{
Dynamic-Shadow~\cite{dynamicshadow} focuses on resolving shadow-object mismatch, addressing inconsistencies between shadows and foreground objects.} Although effective in constrained scenarios, these methods struggle to generalize to long videos or large spatial extrapolation.

To extend applicability, M3DDM~\cite{m3ddm} introduces keyframe-based generation for long-sequence outpainting, whereas Infinite-Canvas~\cite{ic} supports large spatial expansion through global positional guidance. \fix{
M3DDM+~\cite{m3ddmplus} and OutDreamer~\cite{outdreamer} further extend video generation to longer temporal sequences, but they do not support substantial spatial extrapolation.} However, each method addresses only one dimension of the problem, either temporal length or spatial extent, leaving the combined challenge unresolved. In contrast, our method provides a unified  framework capable of handling both long-duration and large spatial expansion ratios for video outpainting within a single model.

\paragraph{Image/Video Inpainting}  
Image inpainting~\cite{irregularholes,generativeinpainting,freeform,repaint,HRIS,realfill} and video inpainting~\cite{deepvideoinpainting,deepflowvideoinpainting,jointtransformer,avid} aim to synthesize missing or new content within user-specified regions of an image or video, and have been widely studied for content editing tasks such as object insertion and object removal. Since these methods are designed to generate arbitrary masked regions, inpainting can be viewed as a generalized form of outpainting. 
However, video outpainting poses greater challenges than inpainting. Specifically, inpainting is conditioned on dense context surrounding the target region, whereas outpainting must generate content beyond a single-sided boundary with no enclosing visual support. In addition, outpainting typically requires synthesizing larger regions in practice. As a result, inpainting-based methods generally fail in outpainting scenarios.

\paragraph{Autoregressive Video Generation}
Another line of work closely related to our setting is autoregressive video generation, which provides a natural mechanism for extending videos to long temporal horizons by conditioning each generation step on previously generated outputs.
This paradigm has therefore been widely explored for long‑range video generation, with some methods adopting LLM‑style token‑based prediction~\cite{vpn,videogpt,phenaki,nuwa} and others applying diffusion models autoregressively across temporal segments~\cite{xie2025progressive,zhang2025generative,huang2025self,streamingt2v,yin2025slow,gao2024ca2}. 
However, these approaches often suffer from error accumulation due to the mismatch between training and inference distributions, where small prediction errors compound over time and progressively degrade visual quality.
In contrast, our method avoids error accumulation by first constructing a \gcg{} that provides global structure, and then generating all frames in parallel rather than sequentially depending on previous predictions.

\begin{figure*}[t]
    \centering
    \includegraphics[width=\linewidth]{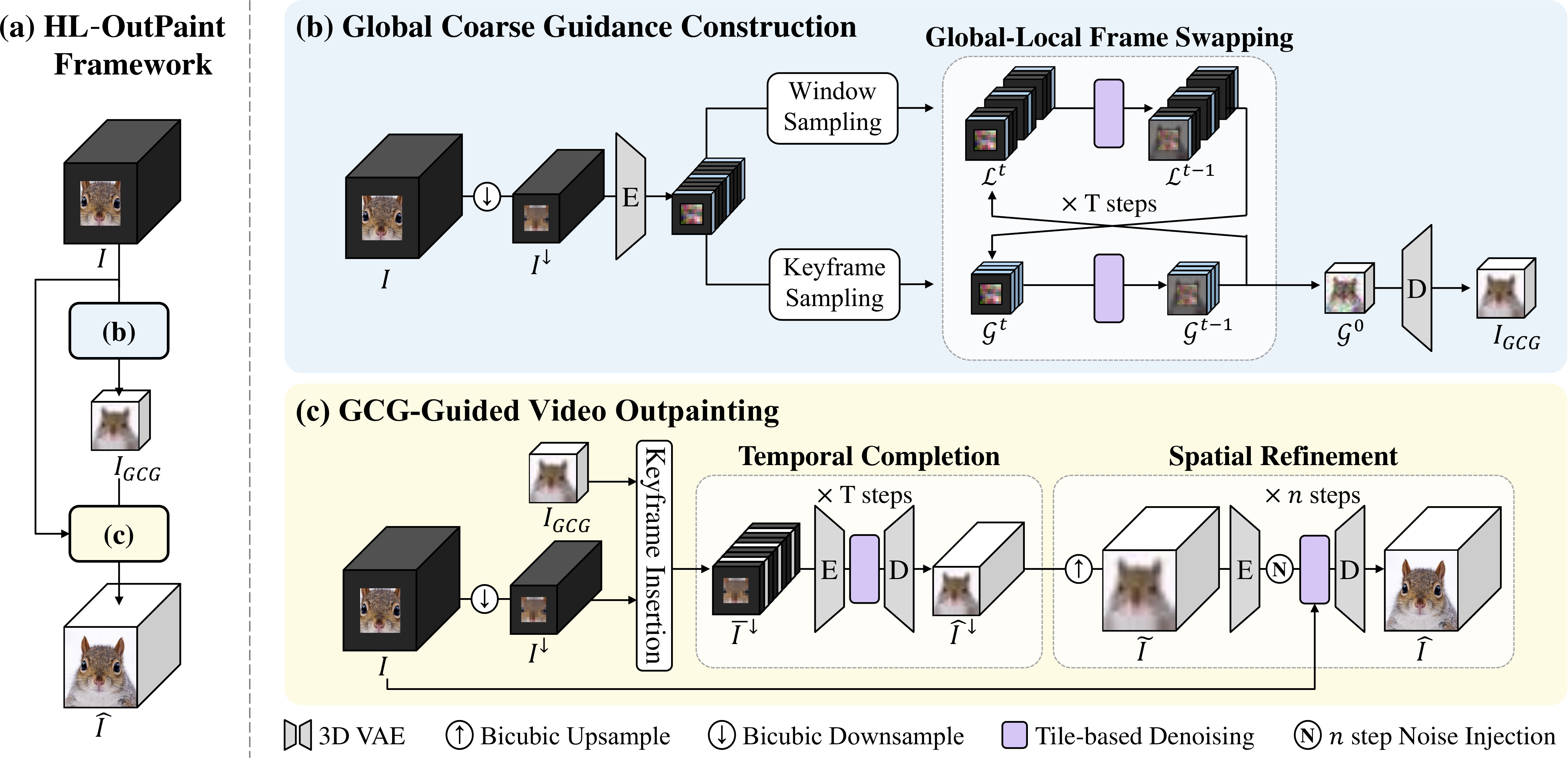}
    \caption{Overall framework of proposed \ours{}. (a) \ours{} consists of two stages: \first{} and \second{}. (b) \first{} generates \gcg{} from spatio-temporally compressed video; at every diffusion timestep t, we perform \swap{} between global keyframes and their local temporal windows to align local and global contexts, producing a globally consistent yet locally well-aligned \gcg{}. (c) \second{} outpaints large-scale video employing \gcg{}.}
    \label{fig:main}
\end{figure*}

\section{Preliminary: Video Outpainting and Diffusion Prior}
\label{sec:pre}

Let $\mathcal{I}' = \{ \mathbf{I}'_f \}_{f=1}^{F}$ denote an original video consisting of $F$ frames, where $\mathbf{I}'_f$ is the $f$-th frame with spatial resolution $H' \times W'$. For video outpainting, the original video is first extended to a larger spatial resolution by appending zero-valued regions beyond the original boundaries, yielding a padded video $\mathcal{I} = \{ \mathbf{I}_f \}_{f=1}^{F}$ with spatial resolution $H \times W$, where $H \ge H'$ and $W \ge W'$. The goal is to synthesize a spatially expanded video $\hat{\mathcal{I}} = \{ \hat{\mathbf{I}}_f \}_{f=1}^{F}$ by generating content in the appended regions. Formally,
\begin{equation}
    \hat{\mathcal{I}} = P(\mathcal{I}, \mathcal{M}),
\end{equation}
where $P(\cdot)$ denotes a video outpainting function and $\mathcal{M}=\{\mathbf{M}_f\}_{f=1}^{F}$ is a set of masks, where $\mathbf{M}_f$ is a binary mask indicating outpainting regions ($1$) and observed regions ($0$) in $\mathbf{I}_f$.

Video diffusion models provide a powerful generative prior for this task, as they can synthesize coherent spatio-temporal content conditioned on partially observed video regions $\mathcal{V}$. For videos with moderate spatial and temporal sizes, one may simply finetune a diffusion model to generate the masked regions conditioned on $\mathcal{V}$ and $\mathcal{M}$. To formalize the denoising process used in such models, we denote by $D$ a diffusion operator that performs a single denoising step in the latent space of a VAE:
\[
    z_{t-1} = D(z_t; \mathcal{V}, \mathcal{M}),
\]
where $z_t$ is the latent representation at timestep $t$. This operator serves as the core generative primitive used throughout our method, including keyframe denoising, local-window refinement, and multi-scale \gcg{} construction.
In typical outpainting settings, the same mask is applied to all input frames, i.e., $\mathbf{M}_f=\mathbf{M}_g$ for all $f,g\in\{1,\dots,F\}$, but in our framework, the masks vary across frames to support keyframe-based conditioning and multi-scale \gcg{} construction.

Existing video diffusion models, however, operate at fixed spatial resolutions and short temporal windows, making them unsuitable for practical high-resolution, long-range video outpainting. Prior work attempts to overcome this limitation using spatio-temporal tiling, but such tiling inevitably introduces inconsistencies across tiles due to limited receptive fields and the absence of global context.

\section{\ours}
\label{sec:method}

\cref{fig:main}(a) illustrates the overall coarse-to-fine framework of \ours{}.
Our framework consists of two stages: (1) \gcg{} construction, which establishes a low-resolution structural backbone for the entire sequence, and (2) \gcg-guided high-resolution outpainting using a tile-based diffusion strategy. 
The \gcg{} construction stage adopts a global-local frame swapping mechanism to ensure both global and local temporal coherence in video outpainting results. For extremely long videos, severe temporal downsampling in \gcg{} construction may break temporal continuity; to handle this, we construct the \gcg{} through a multi-scale iterative refinement process.
In the following subsections, we describe each stage in detail.


\subsection{GCG Construction with Global-Local Frame Swapping}
\label{sec:stage1}

The first stage takes the padded video $\mathcal{I}$ and mask $\mathcal{M}$ as input and constructs a \gcg{} that captures global spatial structure, long-range temporal coherence, and local temporal structure. As illustrated in \cref{fig:main}(b), we begin by spatially downsampling the padded input video $\mathcal{I}$ and mask $\mathcal{M}$ to the resolution supported by the diffusion backbone, obtaining $\mathcal{I}^{\downarrow}$ and $\mathcal{M}^{\downarrow}$. 
We then uniformly sample $K$ keyframes, where $K$ is the maximum number of frames the diffusion model can process in a single forward pass. We denote their indices by $\mathcal{K}=\{k_i\}_{i=1}^{K}$, and the keyframe set by $\mathcal{G}=\{ \mathbf{I}_{k}^{\downarrow} \}_{k \in \mathcal{K}}$.

Each keyframe serves as a temporal anchor that captures global structure across the entire video, but aggressive temporal downsampling inevitably removes fine-scale temporal dynamics. To preserve such local temporal structure, we construct a local temporal window  $\mathcal{L}_i$ for every keyframe $\mathbf{I}^{\downarrow}_{k_i}$. 
Each window contains $K$ frames selected around $k_i$.
These frames are sampled with a small temporal stride $\delta$.
These windows retain short-range temporal cues that are not captured by the keyframes alone. Although the windows themselves are not used as final outputs, they act as auxiliary trajectories that inject fine-scale temporal information into the keyframes during sampling, enabling the keyframes to recover temporal details that would otherwise be lost.

For \gcg{} construction, we initialize latent representations for the keyframe set {$\mathcal{G}$} and for each local temporal window $\mathcal{L}_i$ by sampling Gaussian noise at the resolution of the diffusion model. 
Using the diffusion operator $D$ introduced in \cref{sec:pre}, we then denoise all trajectories in parallel. 
This parallel denoising allows each trajectory to specialize in modeling different aspects of the video: 
the keyframes focus on global structure, while the local windows capture short-range temporal dynamics.

Denoising these trajectories independently would lead to inconsistencies: 
the keyframes would lack fine-scale temporal cues, and the local windows would lack global context. 
To couple their denoising trajectories, we introduce a \swap{} strategy. During the early denoising steps, after each iteration, we replace the latent representation of each frame in the keyframe set with the latent representation of the same frame within its corresponding local temporal window. 
This swapping allows global structure from the keyframes and fine-scale temporal cues from the local windows to be shared and propagated through subsequent denoising steps.

After completing the denoising process with frame \swap{}, we obtain a set of keyframes that are jointly consistent with both global and local temporal structure. These frames form the \gcg{}
$\mathcal{I}_{\mathrm{\gcg}} = \{ \hat{\mathbf{I}}_{k_i}^{\downarrow} \}_{k_i \in \mathcal{K}}$,
which provides stable temporal anchors for the high-resolution outpainting stage.

\paragraph{Multi-scale \gcg{} Construction}
For very long videos, the initial keyframe set {$\mathcal{G}$} may be too sparse to provide sufficient temporal coverage. 
To address this, we adopt a coarse-to-fine, multi-scale guidance construction strategy. 
We first apply the \gcg{} construction process to the uniformly sampled keyframes described above. 
We then refine the guidance by inserting additional keyframes at the temporal midpoints between existing keyframes, forming a denser keyframe set. 
Using the previously constructed guidance as initialization, we reapply the same \gcg{} construction procedure at this finer temporal scale. 
For keyframes that have already been constructed, we set their masks to 1 over the entire frame so that newly inserted keyframes can be outpainted according to the existing ones, while leaving the existing keyframes unchanged. 
This iterative refinement continues until the temporal spacing between adjacent keyframes falls below a predefined threshold $\tau$, yielding a multi-scale guidance that captures both global temporal structure and fine-grained temporal continuity across the entire video.

\paragraph{Adapting the Diffusion Model for \gcg{} Construction}
The \gcg{} construction stage requires a diffusion model capable of synthesizing outpainting regions even when adjacent conditioning frames are far apart in time. Standard video diffusion models are typically trained on densely sampled videos and therefore struggle to synthesize content when the available temporal context is sparse. To adapt the model to this regime, we finetune a DiT-based video diffusion model~\cite{wan2025} on training pairs that mimic the keyframe–window structure used in our guidance construction stage. The video frames are encoded into the latent space of a VAE~\cite{wan2025}, the mask is downsampled accordingly, and both are provided as conditioning signals to the diffusion transformer. This finetuning enables the model to reliably synthesize missing regions under large temporal gaps while maintaining spatial and temporal coherence. Additional architectural and training details are provided in the supplementary document.

\subsection{\gcg-guided Video Outpainting}
\label{sec:stage2}

Given the guidance $\mathcal{I}_{\mathrm{\gcg}}$, we perform \gcg-guided video outpainting to generate the final result $\hat{\mathcal{I}}$. 
\cref{fig:main}(c) illustrates this process. 
As shown in the figure, the input video is outpainted in a coarse-to-fine manner. 
We first perform temporal completion at a reduced spatial resolution, synthesizing the missing regions along the time dimension under the guidance of $\mathcal{I}_{\mathrm{\gcg}}$. 
We then perform spatial refinement, restoring each frame to the original resolution while preserving the temporal consistency established in the coarse stage. 
This two-step process enables high-resolution, long-range video outpainting while maintaining both global structure and fine-scale temporal coherence.

For temporal completion, we construct a low-resolution video whose frames are replaced with corresponding guidance frames when available. 
Specifically, we construct a video $\bar{\mathcal{I}}^{\downarrow} = \{ \bar{\mathbf{I}}_f^{\downarrow} \}_{f=1}^{F}$ and a mask $\bar{\mathcal{M}}^{\downarrow}=\{\bar{\mathbf{M}}_f^{\downarrow}\}_{f=1}^F$ as:
\begin{equation}
(\bar{\mathbf{I}}_f^{\downarrow},\bar{\mathbf{M}}_f^{\downarrow}) =
\begin{cases}
(\hat{\mathbf{I}}_{k_i}^{\downarrow},\mathbf{1}) & \text{if } f = k_i,\; k_i \in \mathcal{K}, \\
(\mathbf{I}_f^{\downarrow},\mathbf{M}_f^{\downarrow}) & \text{otherwise},
\end{cases}
\end{equation}
where $\mathbf{1}$ is an all-one mask. 
We then perform video outpainting to complete the missing regions in $\bar{\mathcal{I}}^{\downarrow}$. 
Since the video length $F$ exceeds the maximum sequence length supported by the diffusion model, we partition the video into overlapping temporal tiles and perform diffusion sampling for them in parallel. 
To ensure smooth transitions between tiles, we blend the overlapping latent regions after every diffusion step. 
For each tile, we apply the video outpainting operator $D$ to synthesize missing regions using the corresponding $\bar{\mathbf{I}}_f^{\downarrow}$ and $\bar{\mathbf{M}}_f^{\downarrow}$ as conditions. 
This process produces a temporally completed low-resolution video $\hat{\mathcal{I}}^{\downarrow}$.

After temporal completion, we perform spatial refinement to recover high-resolution details. 
We first upsample the temporally completed video $\hat{\mathcal{I}}^{\downarrow}$ to the original spatial resolution using bicubic interpolation, yielding an intermediate video $\tilde{\mathcal{I}}$. 
We then regenerate high-frequency content by applying diffusion-based generation in the style of SDEdit~\cite{sdedit}: 
a moderate amount of Gaussian noise is injected into $\tilde{\mathcal{I}}$, and we denoise from an intermediate diffusion step using the outpainting operator $D$. 
During this denoising process, the padded input video $\mathcal{I}$ and mask $\mathcal{M}$ serve as conditioning signals, ensuring that the synthesized regions remain consistent with the observed content.

Because the video resolution and length exceed the capacity of the diffusion model, we apply a spatio-temporal tiling strategy with overlapping tiles. 
Each tile is processed independently by $D$, and at every denoising iteration we blend the overlapping regions of adjacent tiles to maintain spatial and temporal consistency across tile boundaries. 
The temporal completion result provides globally coherent structure for the entire sequence, enabling the tiled refinement to produce a spatially expanded video with consistent long-range temporal continuity. 
This yields the final outpainted video $\hat{\mathcal{I}}$.

\begin{table*}[t]
\caption{
Quantitative comparisons with previous video outpainting methods on the DAVIS~\cite{davis}, DAVIS-20, \fix{YouTube-VOS}~\cite{youtubevos}, Long-Video, and Short-Form datasets, along with a user study.
Input--output resolutions are denoted using arrow notation to indicate spatial extrapolation.
The best and second-best scores are marked in \textbf{bold} and \underline{underline}, respectively.
}
\scalebox{0.88}{
\begin{tabular}{c|c c l ccccccc}
\hline
Video Length & Dataset & Resolution & Method
& PSNR$\uparrow$ & SSIM$\uparrow$ & FVD$\downarrow$
& SC$\uparrow$ & BC$\uparrow$ & AQ$\uparrow$ \\
\hline\hline

\multirow{15}{*}{\makecell[c]{Short \\ (Avg.~68 frames)}}
& \multirow{5}{*}{DAVIS}
& \multirow{5}{*}{\makecell[c]{512$\times$512 \\ $\downarrow$ \\ 1280$\times$720}}
& M3DDM~\cite{m3ddm}        & 15.07 & \underline{0.618} & 514.2  & 0.851 & 0.902 & 0.466 \\
& & & MOTIA~\cite{motia}        & 14.64 & 0.487 & 871.4  & 0.851 & 0.896 & 0.449 \\
& & & Infinite-Canvas~\cite{ic} & 15.85 & 0.590 & 281.8  & 0.872 & 0.909 & 0.489 \\
& & & VACE~\cite{vace}          & \underline{16.63} & 0.618 & \textbf{177.0}
                              & \underline{0.889} & \textbf{0.915} & \underline{0.499} \\
& & & \cellcolor{yellow!30}\ours{}
& \cellcolor{yellow!30}\textbf{16.87}
& \cellcolor{yellow!30}\textbf{0.619}
& \cellcolor{yellow!30}\underline{202.4}
& \cellcolor{yellow!30}\textbf{0.890}
& \cellcolor{yellow!30}\underline{0.910}
& \cellcolor{yellow!30}\textbf{0.502} \\
\cline{2-10}

& \multirow{5}{*}{DAVIS-20}
& \multirow{5}{*}{\makecell[c]{512$\times$512 \\ $\downarrow$ \\ 1920$\times$1080}}
& M3DDM~\cite{m3ddm}        & 11.75 & 0.552 & 2128.9 & 0.770 & 0.862 & 0.409 \\
& & & MOTIA~\cite{motia}        & 13.29 & 0.468 & 2244.9 & 0.826 & 0.884 & 0.422 \\
& & & Infinite-Canvas~\cite{ic} & 13.87 & 0.562 & 1008.4 & 0.839 & 0.892 & 0.487 \\
& & & VACE~\cite{vace}          & \underline{14.89} & \underline{0.591} & \underline{620.0}
                              & \underline{0.875} & \underline{0.897} & \textbf{0.528} \\
& & & \cellcolor{yellow!30}\ours{}
& \cellcolor{yellow!30}\textbf{15.32}
& \cellcolor{yellow!30}\textbf{0.620}
& \cellcolor{yellow!30}\textbf{564.6}
& \cellcolor{yellow!30}\textbf{0.877}
& \cellcolor{yellow!30}\textbf{0.901}
& \cellcolor{yellow!30}\underline{0.520} \\
\cline{2-10}

& \multirow{5}{*}{\fix{YouTube-VOS}}
& \multirow{5}{*}{\makecell[c]{256$\times$256 \\ $\downarrow$ \\ 512$\times$512}}
& M3DDM~\cite{m3ddm}        & 16.06 & \underline{0.575} & \underline{1714.6} & 0.834 & 0.904 & 0.403 \\
& & & MOTIA~\cite{motia}        & 16.11 & 0.526 & 1841.0 & 0.830 & 0.897 & 0.413 \\
& & & Infinite-Canvas~\cite{ic} & \underline{16.71} & \underline{0.592} & 1401.0 & 0.840 & 0.908 & \underline{0.422} \\
& & & VACE~\cite{vace}          & 14.61 & 0.534 & 1753.0 & \textbf{0.869} & \underline{0.919} & 0.413 \\
& & & \cellcolor{yellow!30}\ours{}
& \cellcolor{yellow!30}\textbf{22.15}
& \cellcolor{yellow!30}\textbf{0.821}
& \cellcolor{yellow!30}\textbf{634.0}
& \cellcolor{yellow!30}\underline{0.862}
& \cellcolor{yellow!30}\textbf{0.923}
& \cellcolor{yellow!30}\textbf{0.523} \\
\hline

\multirow{10}{*}{\makecell[c]{Long \\ (Avg.~481 frames)}}
& \multirow{5}{*}{Long-Video}
& \multirow{5}{*}{\makecell[c]{512$\times$512 \\ $\downarrow$ \\ 1280$\times$720}}
& M3DDM~\cite{m3ddm}        & 14.65 & \underline{0.630} & 454.5  & 0.866 & 0.906 & 0.520 \\
& & & MOTIA~\cite{motia}        & 13.99 & 0.471 & 1431.8 & 0.858 & 0.899 & 0.502 \\
& & & Infinite-Canvas~\cite{ic} & 15.31 & 0.586 & 275.0  & 0.869 & 0.906 & 0.532 \\
& & & VACE~\cite{vace}          & \underline{15.84} & 0.620 & \textbf{131.7}
                              & \underline{0.888} & \underline{0.912} & \underline{0.536} \\
& & & \cellcolor{yellow!30}\ours{}
& \cellcolor{yellow!30}\textbf{16.60}
& \cellcolor{yellow!30}\textbf{0.633}
& \cellcolor{yellow!30}\underline{133.2}
& \cellcolor{yellow!30}\textbf{0.889}
& \cellcolor{yellow!30}\textbf{0.912}
& \cellcolor{yellow!30}\textbf{0.555} \\
\cline{2-10}

& \multirow{5}{*}{Short-Form}
& \multirow{5}{*}{\makecell[c]{416$\times$720 \\ $\downarrow$ \\ 1280$\times$720}}
& M3DDM~\cite{m3ddm}        & - & - & - & 0.880 & \underline{0.921} & 0.514 \\
& & & MOTIA~\cite{motia}        & - & - & - & 0.872 & 0.907 & 0.535 \\
& & & Infinite-Canvas~\cite{ic} & - & - & - & 0.878 & 0.912 & \underline{0.569} \\
& & & VACE~\cite{vace}          & - & - & - & \underline{0.900} & 0.911 & 0.550 \\
& & & \cellcolor{yellow!30}\ours{}
& \cellcolor{yellow!30}-
& \cellcolor{yellow!30}-
& \cellcolor{yellow!30}-
& \cellcolor{yellow!30}\textbf{0.920}
& \cellcolor{yellow!30}\textbf{0.930}
& \cellcolor{yellow!30}\textbf{0.574} \\
\hline

\end{tabular}
}
\label{tab:quant}
\end{table*}

\paragraph{Adapting the Diffusion Model for \gcg-guided Video Outpainting}
While we use the outpainting operator $D$ in both the \gcg{} construction stage and the \gcg-guided video outpainting stage, these two stages require diffusion models that operate under different temporal regimes. 
The \gcg{} construction stage must handle sparsely sampled frames, whereas the \gcg-guided video outpainting stage relies on densely sampled frames at the original frame rate. 
To reflect this difference, we finetune a separate diffusion operator $D$ on full-frame-rate training pairs for the second stage. 
This finetuning enables the model to synthesize missing regions under dense temporal conditioning and to support both the low-resolution temporal completion and the subsequent high-resolution spatial refinement.


\section{Experiments}


\paragraph{Implementation Details}
We fine-tune a state-of-the-art video diffusion model, Wan2.2-14B-I2V~\cite{wan2025}, using LoRA~\cite{lora}.
We train two separate LoRA modules for \gcg{} construction and \gcg-guided video outpainting, respectively.
All LoRA modules are trained with a learning rate of $1 \times 10^{-4}$ and a rank of 128 on the OpenVid-1M~\cite{openvid} and REDS~\cite{reds} datasets, where all videos are resampled to a resolution of $768 \times 768$ with 49 frames.
During training, the input video and binary mask are concatenated to the noise latent along the channel dimension.
The two LoRA modules adopt different data sampling strategies: for \gcg{} construction, frame intervals are randomly sampled with large gaps to model sparse keyframe outpainting, whereas \gcg-guided video outpainting uses standard dense temporal sampling.
We empirically set the stride $\delta$  based on the degree of motion in the input video, using $\delta=5$ for static videos and $\delta=1$ for dynamic ones. We set the threshold $\tau$ for multi‑scale \gcg{} construction to 20, which determines when the refinement process terminates.
Additional details are provided in the supplementary document.

\begin{figure*}[t]
  \centering
  \includegraphics[width=\linewidth]{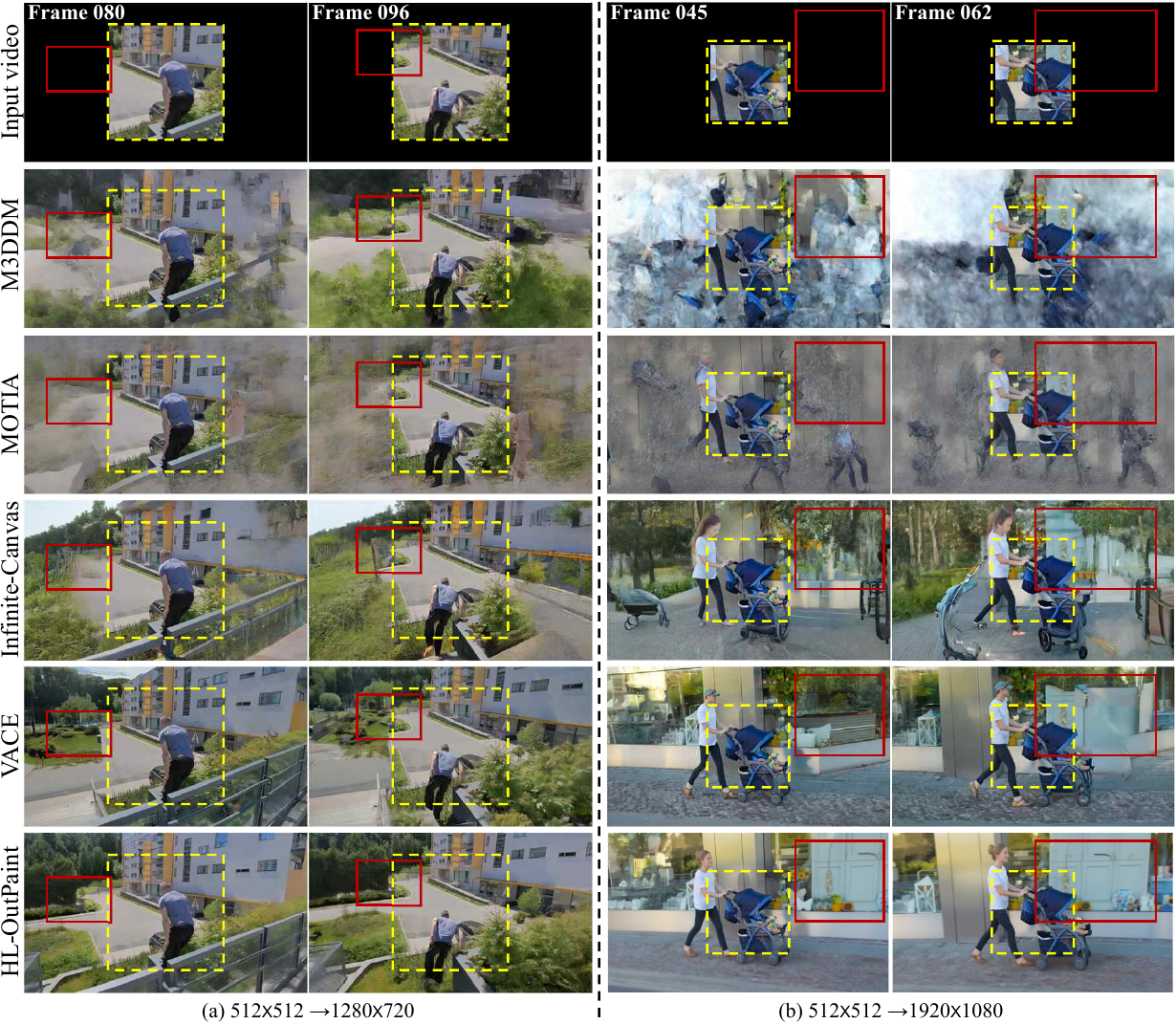}
  \caption{Qualitative comparison on the DAVIS~\cite{davis} dataset with outpainting expansions of (a) $512\times512 \rightarrow 1280\times720$ and (b) $512\times512 \rightarrow 1920\times1080$. The yellow dashed box marks the original region before outpainting. 
  The red box highlights regions where competing methods fail while our method produce coherent results.
  }
  \label{fig:qual_davis}
\end{figure*}
\begin{figure*}[t]
  \centering
  \includegraphics[width=\linewidth]{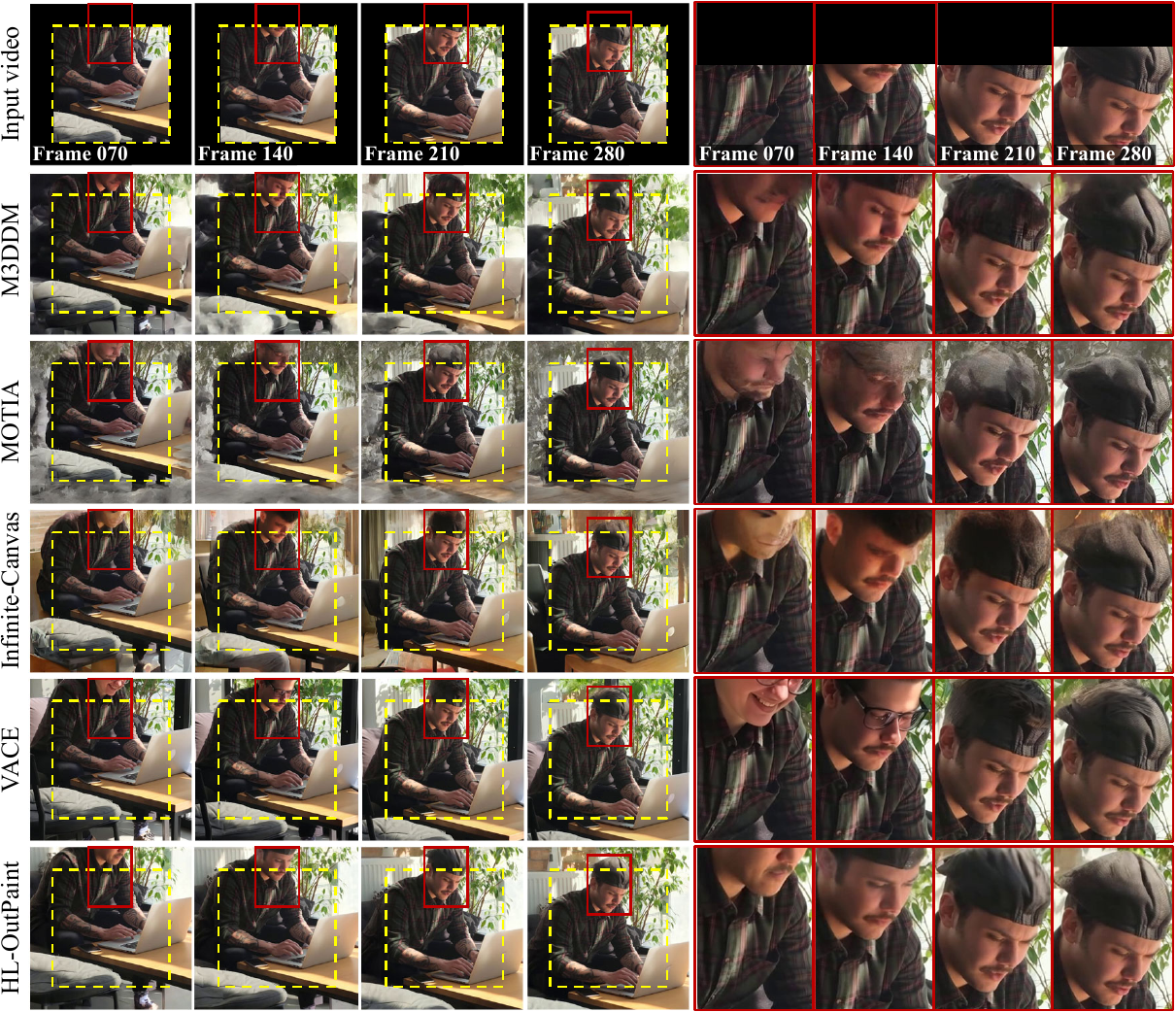}
  \caption{Qualitative comparison on the Long-Video dataset with outpainting expansion of $512\times512 \rightarrow 1280\times720$. The yellow dashed box marks the original region before outpainting. For better visualization, we center-crop the outpainted results to a $720\times720$ region.
  }
  \label{fig:qual_longvideo}
\end{figure*}
\begin{figure*}[t]
  \centering
  \includegraphics[width=\linewidth]{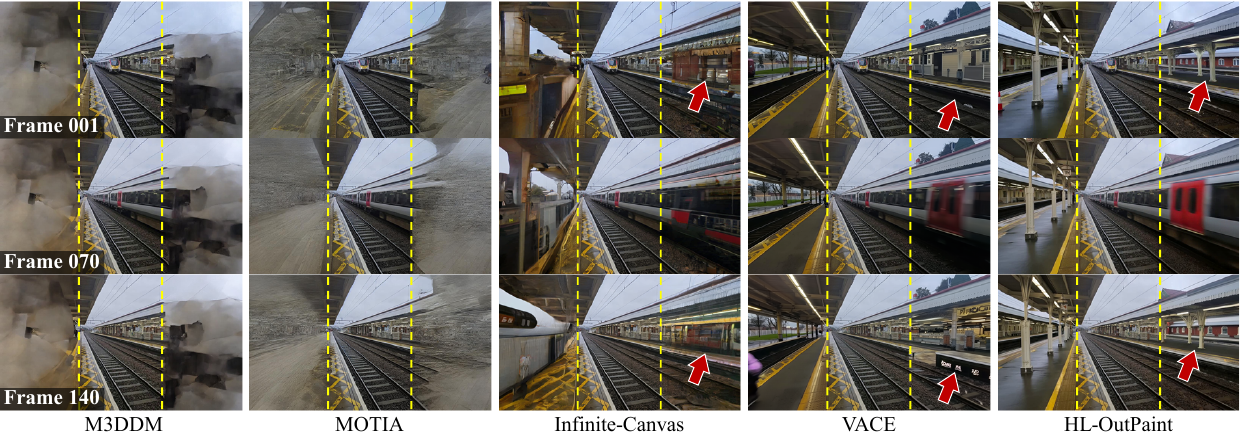}
  \caption{
  Qualitative comparison on the Short-Form dataset with an outpainting expansion of $416 \times 720 \rightarrow 1280 \times 720$. The yellow dashed box denotes the original region before outpainting. The red arrow highlights regions where competing methods fail to maintain long-term temporal coherence, while our method produces stable results.
  }
  \label{fig:qual_shortform}
\end{figure*}
\subsection{Baseline Comparisons}

To demonstrate the effectiveness of \ours{}, we compare our method with various representative video outpainting approaches, including M3DDM~\cite{m3ddm}, MOTIA~\cite{motia}, Infinite-Canvas~\cite{ic}, and VACE~\cite{vace}, \fix{which are built on different generative priors: the first three are based on Stable Diffusion~\cite{HRIS}, while VACE leverages a video diffusion transformer, Wan-2.1~\cite{wan2025}, as its generative prior.} For long-video inference, we follow the inference strategies specified in the original papers or official implementations: MOTIA and Infinite-Canvas employ tile-based generation similar to ours, while VACE adopts an autoregressive scheme that uses the last generated frame as the initial frame for subsequent generation.

For comprehensive evaluation, we conduct experiments on multiple datasets with varying video lengths and spatial expansion scales. We use the DAVIS 2017 training and validation set~\cite{davis}, which contains 90 videos with lengths ranging from 25 to 104 frames, and evaluate spatial extrapolation from $512 \times 512$ to $1280 \times 720$.To assess performance under more challenging spatial extrapolation, we further construct DAVIS-20, a subset of 20 videos randomly sampled from the same dataset, and evaluate a larger expansion from $512 \times 512$ to $1920 \times 1080$.
To further evaluate practical long-video scenarios, we construct two datasets, Long-Video and Short-Form, collected from Pexels\footnote{https://www.pexels.com}, each containing videos of approximately 500 frames.
The Long-Video dataset includes 20 videos and is evaluated under spatial extrapolation from $512 \times 512$ to $1280 \times 720$. In contrast, the Short-Form dataset consists of 9 portrait-format videos and is used to evaluate the conversion from $416 \times 720$ vertical videos to a $1280 \times 720$ horizontal format.

To measure visual fidelity, we use PSNR, SSIM, and Fr\'echet Video Distance (FVD)~\cite{fvd}, following prior works~\cite{m3ddm, ic}.
We also adopt Subject Consistency (SC) and Background Consistency (BC)~\cite{vbench}, which quantify temporal coherence by measuring feature similarity between each frame and both the first frame and its adjacent frames, using DINO~\cite{dino} and CLIP~\cite{clip} features, respectively.
In addition, we report Aesthetic Quality (AQ)~\cite{vbench}, which measures per-frame aesthetic quality using a CLIP-based aesthetic estimator~\cite{clip-a}.

\cref{fig:qual_davis,fig:qual_shortform,fig:qual_longvideo} show qualitative comparisons with baseline methods on the DAVIS~\cite{davis}, Short-Form, and Long-Video datasets, respectively. As shown in these figures, MOTIA~\cite{motia} and M3DDM~\cite{m3ddm} suffer from severe visual artifacts in most cases under large spatial extrapolation. While Infinite-Canvas~\cite{ic} and VACE~\cite{vace} can generate plausible results for individual frames, they often fail to preserve long-term temporal coherence over extended sequences. 
For example, \cref{fig:qual_shortform} presents a video in which a train passes through a platform, temporarily occluding and then revealing the same region.
As highlighted by the red arrow, Infinite-Canvas and VACE produce inconsistent appearances of the same area before and after the occlusion, indicating a failure to maintain temporal coherence. In contrast, our method successfully preserves long-term consistency across such challenging scenarios. 
Quantitative comparisons in \cref{tab:quant} further support these observations, where \ours{} achieves the best performance across most evaluation metrics, demonstrating its overall effectiveness.

\begin{figure}[t]
    \centering
    \includegraphics[width=1.0\linewidth]{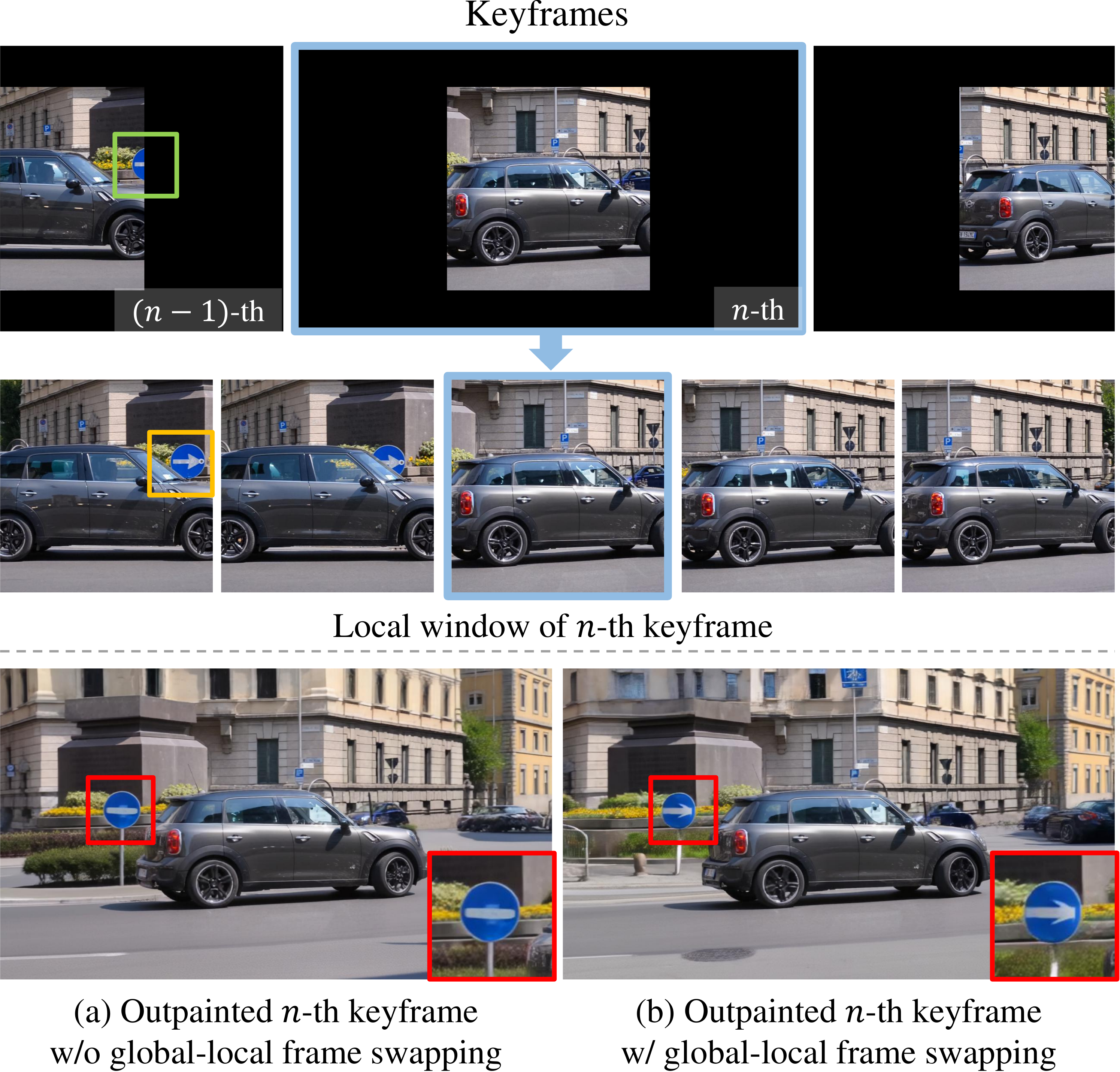}
    \caption{
    Sparse keyframes and the local temporal window centered at the $n$-th keyframe (Top).
    Outpainted $n$-th keyframe without (left) and with (right) \swap{}, highlighting how local window information resolves structural inconsistencies in the keyframe (Bottom). The input videos are from the DAVIS dataset~\cite{davis} (car-roundabout).
    }
    \label{fig:keyframe_localwindow}
\end{figure}
\begin{table}[H]
\caption{
Quantitative comparison with and without \swap{} on the Long-Video dataset. Best results are shown in \textbf{bold}.
}
\scalebox{0.80}{
\begin{tabular}{c|cccccc}
\hline
Global-Local \\ Frame Swapping   & PSNR$\uparrow$   & SSIM$\uparrow$     & FVD$\downarrow$                   & SC$\uparrow$      & BC$\uparrow$      & AQ$\uparrow$      \\\hline\hline
\ding{55}           & {16.21}& {0.6304} & {141.8}     & {0.8872}& {0.9109}& {0.5513}\\
\rowcolor{yellow!30}
\ding{51}           & \textbf{16.60}   & \textbf{0.6332}    & \textbf{133.2}        & \textbf{0.8887}   & \textbf{0.9122}   & \textbf{0.5549}   \\ \hline
\end{tabular}
}
\label{tab:ablation_swapping}
\end{table}

\subsection{Ablation \& Analysis}
\paragraph{Effect of \swap{}}
\cref{fig:keyframe_localwindow} illustrates the problem that \swap{} is designed to address.
Specifically, the green box in the $(n\!-\!1)$-th  keyframe shows a traffic sign that is partially cropped. Since keyframes are sparsely sampled, the subsequent $n$-th keyframe also contains no observation of this sign. 
When keyframes are outpainted without \swap{}, the model must hallucinate the missing region and generates an arbitrary sign shape, as shown in \cref{fig:keyframe_localwindow}(a). 
However, as indicated by the yellow box, neighboring frames already contain a clear arrow-shaped observation. This mismatch between local observations and the hallucinated content leads to severe temporal inconsistency in the outpainting results.
By contrast, \swap{} injects local window information into the \gcg{} construction process, enabling the global keyframe to inherit correct structural cues from nearby frames, as shown in \cref{fig:keyframe_localwindow}(b).
Consequently, \cref{tab:ablation_swapping} shows consistent improvements across all metrics, demonstrating the effectiveness of \swap{}.

\begin{figure}[t]
    \centering
    \includegraphics[width=1.0\linewidth]{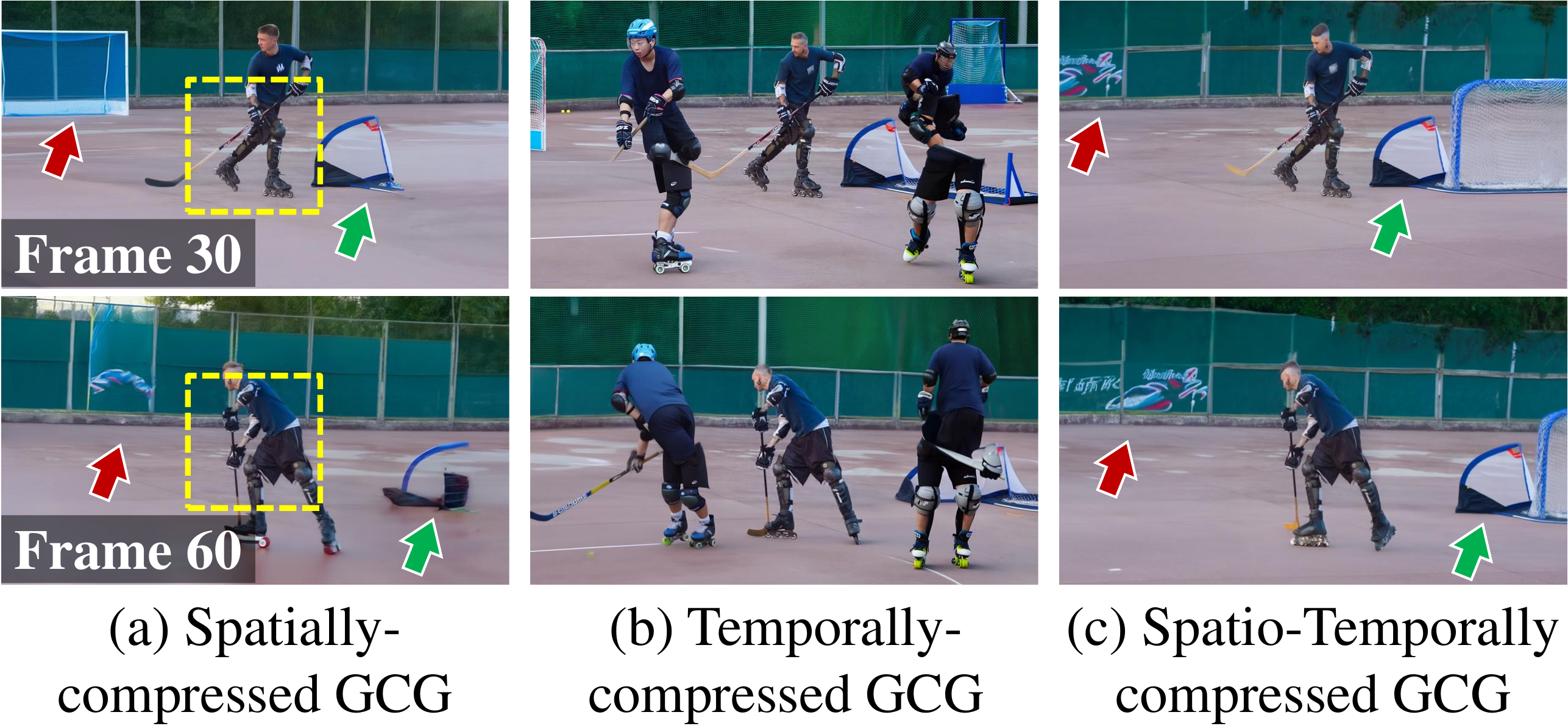}
    \caption{Outpainting results using \gcg{} compressed along different spatial and temporal axes.
Yellow boxes denote the original video region. The input videos are from the DAVIS dataset~\cite{davis} (hockey).
    }
    \label{fig:gcg_ablation}
\end{figure}

\paragraph{Effect of Spatial and Temporal Compression in \gcg}
\cref{fig:gcg_ablation} analyzes the effects of compressing the \gcg{} along the spatial axis only, the temporal axis only, and both axes jointly. 
Without temporal compression, temporal tiles are generated without information exchange, leading to long-term temporal incoherence. 
For example, as illustrated by the red arrows in \cref{fig:gcg_ablation}(a), some objects, such as the goal post, gradually disappear over time.
In contrast, without spatial compression, tiles are generated spatially independently, resulting in severe repetition artifacts across adjacent regions, as shown in \cref{fig:gcg_ablation}(b). Overall, jointly compressing the \gcg{} along both spatial and temporal axes yields the most stable and coherent outpainting results. Detailed experimental settings are provided in the supplementary document.



\begin{figure*}[t]
    \centering
    \includegraphics[width=0.95\linewidth]{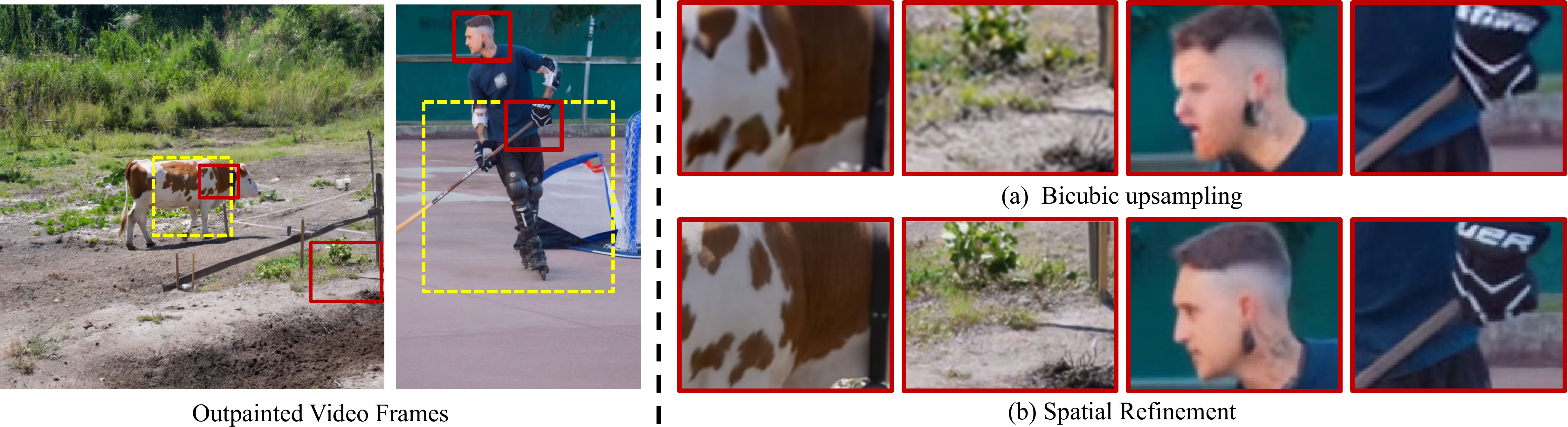}
    \caption{Qualitative comparison between (a) bicubic upsampling of the temporally completed low-resolution video $\hat{\mathcal{I}}^{\downarrow}$ and (b) spatial refinement results applied to the bicubic-upsampled video. The input videos are from the DAVIS dataset~\cite{davis} (cow, hockey).}
    \label{fig:ablation_sr}
\end{figure*}

\paragraph{Effect of Spatial Refinement}
As shown in \cref{fig:ablation_sr}(a), bicubic upsampling of the temporally completed low-resolution video $\hat{\mathcal{I}}^{\downarrow}$ results in blurry details in both the original input and the outpainted regions.
By applying spatial refinement to the blurred upsampled video, we restore high-frequency details and improve structural fidelity.
For example, as shown in \cref{fig:ablation_sr}(b), applying spatial refinement produces more natural and sharper facial contours and features, while also restoring the shapes of characters in text regions more clearly. Moreover, in regions with complex textures, such as grass and cow fur, fine-grained patterns are better preserved and enhanced, resulting in substantially more realistic visual quality compared to simple upsampling. This demonstrates that the proposed spatial refinement stage plays an essential role in transforming the low-resolution temporally completed results into high-resolution videos, effectively contributing to detail restoration and improved visual realism in the final output.



\section{Conclusion}
In this paper, we proposed \ours{}, a novel video outpainting method designed to support large spatial extrapolation over long sequences while preserving spatio-temporal coherence. Our coarse-to-fine framework first constructs a \gcg{} to capture global structure and motion, which is then refined into spatially detailed and temporally consistent high-resolution results. \fix{To ensure both global and local temporal coherence, we introduced a global-local frame swapping mechanism.} Experimental results demonstrate the effectiveness of \ours{} across diverse and challenging scenarios, particularly in maintaining stability during large spatial expansions over extended video sequences.

\begin{figure}[t]
    \centering
    \includegraphics[width=1.0\linewidth]{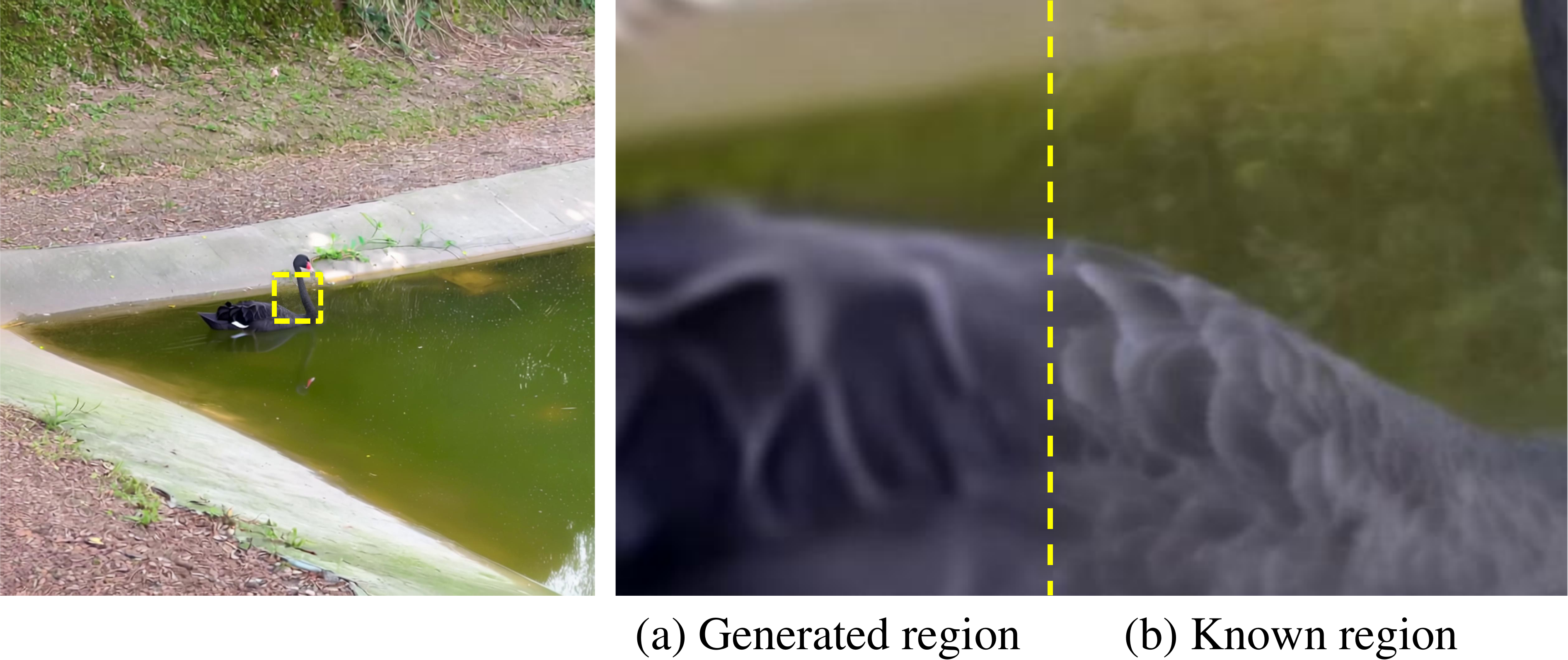}
    \caption{\fix{
    Failure case under extreme spatial expansion (512$\times$512 $\rightarrow$ 5760$\times$5760).
    The input is heavily downsampled (e.g., to 768$\times$768) during \gcg{} construction, causing significant loss of high-frequency details.
    While the original regions are restored during refinement due to strong conditioning, the outpainted regions fail to recover fine details, resulting in blurry structures. The input videos are from the DAVIS dataset~\cite{davis} (black swan).
    }}
    \label{fig:limitation_extreme}
\end{figure}

\paragraph{Limitations}
\ours{} is not suitable for real-time applications, as it generates all frames jointly in a single inference process.
In addition, our method may fail in extremely large or long cases.
For extremely large spatial expansion, the input video must be heavily downsampled to the resolution supported by the diffusion model for \gcg{} construction, which can cause critical information loss.
As shown in Fig.~\ref{fig:limitation_extreme}, when expanding a video from 512$\times$512 to 5760$\times$5760, the input may need to be downsampled to 768$\times$768 to construct the \gcg{}.
Although the original regions can recover fine details during the refinement stage due to strong conditioning from the input frames, the outpainted regions rely solely on the \gcg{}.
As a result, high-frequency details lost during \gcg{} construction cannot be effectively recovered during upsampling and refinement, often leading to overly smooth or blurry synthesized regions.
For very long videos, keyframes and local windows may also fail to fully cover the entire sequence, making it difficult to maintain temporal consistency.
However, such extreme cases are rarely encountered in practical video outpainting scenarios.




\begin{acks} 
This work was supported by Samsung Electronics Co., Ltd.; by the Institute of Information \& Communications Technology Planning \& Evaluation (IITP) grant funded by the Korea government (MSIT) (Artificial Intelligence Graduate School Program (POSTECH), No. RS-2019-II191906); and by the National Research Foundation of Korea (NRF) grant funded by the Korea government (MSIT) (No. RS-2026-25492695).
\end{acks}

\bibliographystyle{ACM-Reference-Format}
\bibliography{sections/reference}

@article{sdedit,
  title={Sdedit: Guided image synthesis and editing with stochastic differential equations},
  author={Meng, Chenlin and He, Yutong and Song, Yang and Song, Jiaming and Wu, Jiajun and Zhu, Jun-Yan and Ermon, Stefano},
  journal={arXiv preprint arXiv:2108.01073},
  year={2021}
}

@misc{clip-a,
  author       = {LAION-AI},
  title        = {Aesthetic Predictor},
  year         = {2023},
  howpublished = {\url{https://github.com/LAION-AI/aesthetic-predictor}},
  note         = {Accessed: 2025-05-01}
}

@inproceedings{clip,
  title={Learning transferable visual models from natural language supervision},
  author={Radford, Alec and Kim, Jong Wook and Hallacy, Chris and Ramesh, Aditya and Goh, Gabriel and Agarwal, Sandhini and Sastry, Girish and Askell, Amanda and Mishkin, Pamela and Clark, Jack and others},
  booktitle={International conference on machine learning},
  pages={8748--8763},
  year={2021},
  organization={PmLR}
}

@article{dino,
  title={Dino: Detr with improved denoising anchor boxes for end-to-end object detection},
  author={Zhang, Hao and Li, Feng and Liu, Shilong and Zhang, Lei and Su, Hang and Zhu, Jun and Ni, Lionel M and Shum, Heung-Yeung},
  journal={arXiv preprint arXiv:2203.03605},
  year={2022}
}

@article{gao2024ca2,
  title={Ca2-vdm: Efficient autoregressive video diffusion model with causal generation and cache sharing},
  author={Gao, Kaifeng and Shi, Jiaxin and Zhang, Hanwang and Wang, Chunping and Xiao, Jun and Chen, Long},
  journal={arXiv preprint arXiv:2411.16375},
  year={2024}
}

@inproceedings{yin2025slow,
  title={From slow bidirectional to fast autoregressive video diffusion models},
  author={Yin, Tianwei and Zhang, Qiang and Zhang, Richard and Freeman, William T and Durand, Fredo and Shechtman, Eli and Huang, Xun},
  booktitle={Proceedings of the Computer Vision and Pattern Recognition Conference},
  pages={22963--22974},
  year={2025}
}

@article{huang2025self,
  title={Self Forcing: Bridging the Train-Test Gap in Autoregressive Video Diffusion},
  author={Huang, Xun and Li, Zhengqi and He, Guande and Zhou, Mingyuan and Shechtman, Eli},
  journal={arXiv preprint arXiv:2506.08009},
  year={2025}
}

@article{zhang2025generative,
  title={Generative pre-trained autoregressive diffusion transformer},
  author={Zhang, Yuan and Jiang, Jiacheng and Ma, Guoqing and Lu, Zhiying and Huang, Haoyang and Yuan, Jianlong and Duan, Nan and Jiang, Daxin},
  journal={arXiv preprint arXiv:2505.07344},
  year={2025}
}

@inproceedings{xie2025progressive,
  title={Progressive autoregressive video diffusion models},
  author={Xie, Desai and Xu, Zhan and Hong, Yicong and Tan, Hao and Liu, Difan and Liu, Feng and Kaufman, Arie and Zhou, Yang},
  booktitle={Proceedings of the Computer Vision and Pattern Recognition Conference},
  pages={6322--6332},
  year={2025}
}

@article{nuwa,
  title={Nuwa-infinity: Autoregressive over autoregressive generation for infinite visual synthesis},
  author={Liang, Jian and Wu, Chenfei and Hu, Xiaowei and Gan, Zhe and Wang, Jianfeng and Wang, Lijuan and Liu, Zicheng and Fang, Yuejian and Duan, Nan},
  journal={Advances in Neural Information Processing Systems},
  volume={35},
  pages={15420--15432},
  year={2022}
}

@inproceedings{vpn,
  title={Video pixel networks},
  author={Kalchbrenner, Nal and Oord, A{\"a}ron and Simonyan, Karen and Danihelka, Ivo and Vinyals, Oriol and Graves, Alex and Kavukcuoglu, Koray},
  booktitle={International Conference on Machine Learning},
  pages={1771--1779},
  year={2017},
  organization={PMLR}
}

@article{videogpt,
  title={Videogpt: Video generation using vq-vae and transformers},
  author={Yan, Wilson and Zhang, Yunzhi and Abbeel, Pieter and Srinivas, Aravind},
  journal={arXiv preprint arXiv:2104.10157},
  year={2021}
}

@article{phenaki,
  title={Phenaki: Variable length video generation from open domain textual description},
  author={Villegas, Ruben and Babaeizadeh, Mohammad and Kindermans, Pieter-Jan and Moraldo, Hernan and Zhang, Han and Saffar, Mohammad Taghi and Castro, Santiago and Kunze, Julius and Erhan, Dumitru},
  journal={arXiv preprint arXiv:2210.02399},
  year={2022}
}

@INPROCEEDINGS{dehan,
    author={Dehan, Loïc and Van Ranst, Wiebe and Vandewalle, Patrick and Goedemé, Toon},
    booktitle={2022 IEEE/CVF Conference on Computer Vision and Pattern Recognition Workshops (CVPRW)},
    title={Complete and temporally consistent video outpainting},
    year={2022},
    volume={},
    number={},
    pages={686-694},
    doi={10.1109/CVPRW56347.2022.00084}
}

@inproceedings{m3ddm,
    title={Hierarchical Masked 3D Diffusion Model for Video Outpainting},
    author={Fan, Fanda and Guo, Chaoxu and Gong, Litong and Wang, Biao and Ge, Tiezheng and Jiang, Yuning and Luo, Chunjie and Zhan, Jianfeng},
    booktitle={Proceedings of the 31st ACM International Conference on Multimedia},
    pages={7890--7900},
    year={2023}
}

@inproceedings{motia,
    author = {Wang, Fu-Yun and Wu, Xiaoshi and Huang, Zhaoyang and Shi, Xiaoyu and Shen, Dazhong and Song, Guanglu and Liu, Yu and Li, Hongsheng},
    title = {Be-Your-Outpainter: Mastering Video Outpainting Through Input-Specific Adaptation},
    year = {2024},
    isbn = {978-3-031-72783-2},
    publisher = {Springer-Verlag},
    address = {Berlin, Heidelberg},
    url = {[https://doi.org/10.1007/978-3-031-72784-9_9](https://doi.org/10.1007/978-3-031-72784-9_9)},
    doi = {10.1007/978-3-031-72784-9_9},
    booktitle = {Computer Vision – ECCV 2024: 18th European Conference, Milan, Italy, September 29–October 4, 2024, Proceedings, Part XLIV},
    pages = {153–168},
    numpages = {16},
    location = {Milan, Italy}
}

@article{ic, 
    title={Infinite-Canvas: Higher-Resolution Video Outpainting with Extensive Content Generation}, 
    volume={39}, 
    url={https://ojs.aaai.org/index.php/AAAI/article/view/32213}, 
    DOI={10.1609/aaai.v39i2.32213}, 
    number={2}, journal={Proceedings of the AAAI Conference on Artificial Intelligence}, 
    author={Chen, Qihua and Ma, Yue and Wang, Hongfa and Yuan, Junkun and Zhao, Wenzhe and Tian, Qi and Wang, Hongmei and Min, Shaobo and Chen, Qifeng and Liu, Wei}, 
    year={2025}, 
    month={Apr.}, 
    pages={2150-2158} 
}

@inproceedings{vace,
    title = {VACE: All-in-One Video Creation and Editing},
    author = {Jiang, Zeyinzi and Han, Zhen and Mao, Chaojie and Zhang, Jingfeng and Pan, Yulin and Liu, Yu},
    booktitle = {Proceedings of the IEEE/CVF International Conference on Computer Vision},
    pages = {17191-17202},
    year = {2025}
}

@INPROCEEDINGS{unboxed,
    author={Yu, Zhongrui and Megaro-Boldini, Martina and Sumner, Robert W. and Djelouah, Abdelaziz},
    booktitle={2025 IEEE/CVF Conference on Computer Vision and Pattern Recognition (CVPR)},
    title={Unboxed: Geometrically and Temporally Consistent Video Outpainting},
    year={2025},
    volume={},
    number={},
    pages={7309-7319},
    doi={10.1109/CVPR52734.2025.00685}
}

@inproceedings{PAO,
    title={Progressive Artwork Outpainting via Latent Diffusion Models},
    author={Song, Dae-Young and Yu, Jung-Jae and Cho, Donghyeon},
    booktitle={Proceedings of the IEEE/CVF International Conference on Computer Vision (ICCV)},
    pages={15405--15415},
    year={2025}
}

@misc{outdreamer,
    title={OutDreamer: Video Outpainting with a Diffusion Transformer},
    author={Linhao Zhong and Fan Li and Yi Huang and Jianzhuang Liu and Renjing Pei and Fenglong Song},
    year={2025},
    eprint={2506.22298},
    archivePrefix={arXiv},
    primaryClass={[cs.CV](http://cs.cv/)},
    url={[https://arxiv.org/abs/2506.22298](https://arxiv.org/abs/2506.22298)},
}

@article{InOut,
  title={InOut: Diverse Image Outpainting via GAN Inversion},
  author={Yen-Chi Cheng and Chieh Hubert Lin and Hsin-Ying Lee and Jian Ren and S. Tulyakov and Ming-Hsuan Yang},
  journal={2022 IEEE/CVF Conference on Computer Vision and Pattern Recognition (CVPR)},
  year={2021},
  pages={11421-11430},
  url={https://api.semanticscholar.org/CorpusID:232478397}
}

@INPROCEEDINGS {SEIO,
    author = { Yu, Hang and Li, Ruilin and Xie, Shaorong and Qiu, Jiayan },
    booktitle = { 2024 IEEE/CVF Conference on Computer Vision and Pattern Recognition (CVPR) },
    title = {{ Shadow-Enlightened Image Outpainting }},
    year = {2024},
    volume = {},
    ISSN = {},
    pages = {7850-7860},
    doi = {10.1109/CVPR52733.2024.00750},
    url = {https://doi.ieeecomputersociety.org/10.1109/CVPR52733.2024.00750},
    publisher = {IEEE Computer Society},
    address = {Los Alamitos, CA, USA},
    month =Jun
}

@inproceedings{CMIO,
    title={Continuous-Multiple Image Outpainting in One-Step via Positional Query and A Diffusion-based Approach},
    author={Shaofeng Zhang and Jinfa Huang and Qiang Zhou and zhibin wang and Fan Wang and Jiebo Luo and Junchi Yan},
    booktitle={The Twelfth International Conference on Learning Representations},
    year={2024},
    url={https://openreview.net/forum?id=7hxoYxKDTV}
}

@article{davis,
  author       = {Jordi Pont{-}Tuset and
                  Federico Perazzi and
                  Sergi Caelles and
                  Pablo Arbel{\'{a}}ez and
                  Alexander Sorkine{-}Hornung and
                  Luc Van Gool},
  title        = {The 2017 {DAVIS} Challenge on Video Object Segmentation},
  journal      = {CoRR},
  volume       = {abs/1704.00675},
  year         = {2017},
  url          = {http://arxiv.org/abs/1704.00675},
  eprinttype    = {arXiv},
  eprint       = {1704.00675},
  bibsource    = {dblp computer science bibliography, https://dblp.org}
}

@inproceedings{youtubevos,
    author = {Xu, Ning and Yang, Linjie and Fan, Yuchen and Yang, Jianchao and Yue, Dingcheng and Liang, Yuchen and Price, Brian and Cohen, Scott and Huang, Thomas},
    title = {YouTube-VOS: Sequence-to-Sequence Video Object Segmentation},
    year = {2018},
    isbn = {978-3-030-01227-4},
    publisher = {Springer-Verlag},
    address = {Berlin, Heidelberg},
    url = {https://doi.org/10.1007/978-3-030-01228-1_36},
    doi = {10.1007/978-3-030-01228-1_36},
    booktitle = {Computer Vision – ECCV 2018: 15th European Conference, Munich, Germany, September 8–14, 2018, Proceedings, Part V},
    pages = {603–619},
    numpages = {17},
    location = {Munich, Germany}
}

@misc{pexels,
  title        = {Pexels},
  author       = {{Pexels}},
  howpublished = {\url{https://www.pexels.com}},
  year         = {2026},
  note         = {Accessed: 2026-01-08}
}

@article{fvd,
  title={Towards Accurate Generative Models of Video: A New Metric \& Challenges},
  author={Thomas Unterthiner and Sjoerd van Steenkiste and Karol Kurach and Rapha{\"e}l Marinier and Marcin Michalski and Sylvain Gelly},
  journal={ArXiv},
  year={2018},
  volume={abs/1812.01717},
  url={https://api.semanticscholar.org/CorpusID:54458806}
}

@ARTICLE{ssim,
  author={Zhou Wang and Bovik, A.C. and Sheikh, H.R. and Simoncelli, E.P.},
  journal={IEEE Transactions on Image Processing}, 
  title={Image quality assessment: from error visibility to structural similarity}, 
  year={2004},
  volume={13},
  number={4},
  pages={600-612},
  doi={10.1109/TIP.2003.819861}
}

@InProceedings{vbench,
     title={{VBench}: Comprehensive Benchmark Suite for Video Generative Models},
     author={Huang, Ziqi and He, Yinan and Yu, Jiashuo and Zhang, Fan and Si, Chenyang and Jiang, Yuming and Zhang, Yuanhan and Wu, Tianxing and Jin, Qingyang and Chanpaisit, Nattapol and Wang, Yaohui and Chen, Xinyuan and Wang, Limin and Lin, Dahua and Qiao, Yu and Liu, Ziwei},
     booktitle={Proceedings of the IEEE/CVF Conference on Computer Vision and Pattern Recognition},
     year={2024}
 }

@inproceedings{lora,
    title={Lo{RA}: Low-Rank Adaptation of Large Language Models},
    author={Edward J Hu and Yelong Shen and Phillip Wallis and Zeyuan Allen-Zhu and Yuanzhi Li and Shean Wang and Lu Wang and Weizhu Chen},
    booktitle={International Conference on Learning Representations},
    year={2022},
    url={https://openreview.net/forum?id=nZeVKeeFYf9},
}

@InProceedings{maskgit,
      title = {MaskGIT: Masked Generative Image Transformer},
      author={Huiwen Chang and Han Zhang and Lu Jiang and Ce Liu and William T. Freeman},
      booktitle = {The IEEE Conference on Computer Vision and Pattern Recognition (CVPR)},
      month = {June},
      year = {2022}
}

@INPROCEEDINGS {EGP,
    author = { Lin, Han and Pagnucco, Maurice and Song, Yang },
    booktitle = { 2021 IEEE/CVF Conference on Computer Vision and Pattern Recognition Workshops (CVPRW) },
    title = {{ Edge Guided Progressively Generative Image Outpainting }},
    year = {2021},
    volume = {},
    ISSN = {},
    pages = {806-815},
    doi = {10.1109/CVPRW53098.2021.00090},
    url = {https://doi.ieeecomputersociety.org/10.1109/CVPRW53098.2021.00090},
    publisher = {IEEE Computer Society},
    address = {Los Alamitos, CA, USA},
    month =Jun
}

@inproceedings{Palette,
      title     = {Palette: Image-to-Image Diffusion Models},
      author    = {Saharia, Chitwan and Chan, William and Chang, Huiwen and Lee, Chris A. and Ho, Jonathan and Salimans, Tim and Fleet, David J. and Norouzi, Mohammad},
      booktitle = {ACM SIGGRAPH 2022 Conference Proceedings},
      year      = {2022},
      pages     = {1--10}
}

@article{wan2025,
      title={Wan: Open and Advanced Large-Scale Video Generative Models}, 
      author={Team Wan and Ang Wang and Baole Ai and Bin Wen and Chaojie Mao and Chen-Wei Xie and Di Chen and Feiwu Yu and Haiming Zhao and Jianxiao Yang and Jianyuan Zeng and Jiayu Wang and Jingfeng Zhang and Jingren Zhou and Jinkai Wang and Jixuan Chen and Kai Zhu and Kang Zhao and Keyu Yan and Lianghua Huang and Mengyang Feng and Ningyi Zhang and Pandeng Li and Pingyu Wu and Ruihang Chu and Ruili Feng and Shiwei Zhang and Siyang Sun and Tao Fang and Tianxing Wang and Tianyi Gui and Tingyu Weng and Tong Shen and Wei Lin and Wei Wang and Wei Wang and Wenmeng Zhou and Wente Wang and Wenting Shen and Wenyuan Yu and Xianzhong Shi and Xiaoming Huang and Xin Xu and Yan Kou and Yangyu Lv and Yifei Li and Yijing Liu and Yiming Wang and Yingya Zhang and Yitong Huang and Yong Li and You Wu and Yu Liu and Yulin Pan and Yun Zheng and Yuntao Hong and Yupeng Shi and Yutong Feng and Zeyinzi Jiang and Zhen Han and Zhi-Fan Wu and Ziyu Liu},
      journal = {arXiv preprint arXiv:2503.20314},
      year={2025}
}

@article{streamingt2v,
  title={StreamingT2V: Consistent, Dynamic, and Extendable Long Video Generation from Text},
  author={Henschel, Roberto and Khachatryan, Levon and Hayrapetyan, Daniil and Poghosyan, Hayk and Tadevosyan, Vahram and Wang, Zhangyang and Navasardyan, Shant and Shi, Humphrey},
  journal={arXiv preprint arXiv:2403.14773},
  year={2024}
}

@inproceedings{bridgingthegap,
    author = {Li, Na and Li, Zihao and Tang, Zuoli and Yu, Yuqing and Zou, Lixin and Li, Chenliang},
    title = {Bridging the Gap: Consistent Image Outpainting via Training-Free Noise Optimization},
    year = {2025},
    isbn = {9798400720352},
    publisher = {Association for Computing Machinery},
    address = {New York, NY, USA},
    url = {https://doi.org/10.1145/3746027.3755278},
    doi = {10.1145/3746027.3755278},
    booktitle = {Proceedings of the 33rd ACM International Conference on Multimedia},
    pages = {9969–9977},
    numpages = {9},
    location = {Dublin, Ireland},
    series = {MM '25}
}

@misc{VIP,
      title={VIP: Versatile Image Outpainting Empowered by Multimodal Large Language Model}, 
      author={Jinze Yang and Haoran Wang and Zining Zhu and Chenglong Liu and Meng Wymond Wu and Mingming Sun},
      year={2024},
      eprint={2406.01059},
      archivePrefix={arXiv},
      primaryClass={cs.CV},
      url={https://arxiv.org/abs/2406.01059}, 
}

@inproceedings{ar,
    author = {Chen, Mark and Radford, Alec and Child, Rewon and Wu, Jeff and Jun, Heewoo and Luan, David and Sutskever, Ilya},
    title = {Generative pretraining from pixels},
    year = {2020},
    publisher = {JMLR.org},
    booktitle = {Proceedings of the 37th International Conference on Machine Learning},
    articleno = {158},
    numpages = {13},
    series = {ICML'20}
}

@inproceedings{pathakCVPR16context,
    Author = {Pathak, Deepak and
    Kr\"ahenb\"uhl, Philipp and
    Donahue, Jeff and
    Darrell, Trevor and
    Efros, Alexei},
    Title = {Context Encoders:
    Feature Learning by Inpainting},
    Booktitle = CVPR,
    Year = {2016}
}

@article{openvid,
  title={OpenVid-1M: A Large-Scale High-Quality Dataset for Text-to-video Generation},
  author={Nan, Kepan and Xie, Rui and Zhou, Penghao and Fan, Tiehan and Yang, Zhenheng and Chen, Zhijie and Li, Xiang and Yang, Jian and Tai, Ying},
  journal={arXiv preprint arXiv:2407.02371},
  year={2024}
}

@InProceedings{reds,
  author = {Nah, Seungjun and Baik, Sungyong and Hong, Seokil and Moon, Gyeongsik and Son, Sanghyun and Timofte, Radu and Lee, Kyoung Mu},
  title = {NTIRE 2019 Challenge on Video Deblurring and Super-Resolution: Dataset and Study},
  booktitle = {CVPR Workshops},
  month = {June},
  year = {2019}
}

@inproceedings{irregularholes,
   author    = {Guilin Liu and Fitsum A. Reda and Kevin J. Shih and Ting-Chun Wang and Andrew Tao and Bryan Catanzaro},
   title     = {Image Inpainting for Irregular Holes Using Partial Convolutions},
   booktitle = {The European Conference on Computer Vision (ECCV)},   
   year      = {2018},
}

@article{generativeinpainting,
  title={Generative Image Inpainting with Contextual Attention},
  author={Yu, Jiahui and Lin, Zhe and Yang, Jimei and Shen, Xiaohui and Lu, Xin and Huang, Thomas S},
  journal={arXiv preprint arXiv:1801.07892},
  year={2018}
}

@misc{freeform,
      title={Free-Form Image Inpainting with Gated Convolution}, 
      author={Jiahui Yu and Zhe Lin and Jimei Yang and Xiaohui Shen and Xin Lu and Thomas Huang},
      year={2019},
      eprint={1806.03589},
      archivePrefix={arXiv},
      primaryClass={cs.CV},
      url={https://arxiv.org/abs/1806.03589}, 
}

@INPROCEEDINGS{repaint,
  author={Lugmayr, Andreas and Danelljan, Martin and Romero, Andres and Yu, Fisher and Timofte, Radu and Van Gool, Luc},
  booktitle={2022 IEEE/CVF Conference on Computer Vision and Pattern Recognition (CVPR)}, 
  title={RePaint: Inpainting using Denoising Diffusion Probabilistic Models}, 
  year={2022},
  volume={},
  number={},
  pages={11451-11461},
  doi={10.1109/CVPR52688.2022.01117}
}

@misc{HRIS,
      title={High-Resolution Image Synthesis with Latent Diffusion Models}, 
      author={Robin Rombach and Andreas Blattmann and Dominik Lorenz and Patrick Esser and Björn Ommer},
      year={2022},
      eprint={2112.10752},
      archivePrefix={arXiv},
      primaryClass={cs.CV},
      url={https://arxiv.org/abs/2112.10752}, 
}

@article{realfill,
    author = {Tang, Luming and Ruiz, Nataniel and Chu, Qinghao and Li, Yuanzhen and Holynski, Aleksander and Jacobs, David E. and Hariharan, Bharath and Pritch, Yael and Wadhwa, Neal and Aberman, Kfir and Rubinstein, Michael},
    title = {RealFill: Reference-Driven Generation for Authentic Image Completion},
    year = {2024},
    issue_date = {July 2024},
    publisher = {Association for Computing Machinery},
    address = {New York, NY, USA},
    volume = {43},
    number = {4},
    issn = {0730-0301},
    url = {https://doi.org/10.1145/3658237},
    doi = {10.1145/3658237},
    journal = {ACM Trans. Graph.},
    month = jul,
    articleno = {135},
    numpages = {12}
}

@INPROCEEDINGS {deepvideoinpainting,
    author = { Kim, Dahun and Woo, Sanghyun and Lee, Joon-Young and Kweon, In So },
    booktitle = { 2019 IEEE/CVF Conference on Computer Vision and Pattern Recognition (CVPR) },
    title = {{ Deep Video Inpainting }},
    year = {2019},
    volume = {},
    ISSN = {},
    pages = {5785-5794},
    doi = {10.1109/CVPR.2019.00594},
    url = {https://doi.ieeecomputersociety.org/10.1109/CVPR.2019.00594},
    publisher = {IEEE Computer Society},
    address = {Los Alamitos, CA, USA},
    month =Jun
}

@InProceedings{deepflowvideoinpainting,
    author = {Xu, Rui and Li, Xiaoxiao and Zhou, Bolei and Loy, Chen Change},
    title = {Deep Flow-Guided Video Inpainting},
    booktitle = {The IEEE Conference on Computer Vision and Pattern Recognition (CVPR)},
    month = {June},
    year = {2019}
}

@inproceedings{jointtransformer,
  author = {Zeng, Yanhong and Fu, Jianlong and Chao, Hongyang},
  title = {Learning Joint Spatial-Temporal Transformations for Video Inpainting},
  booktitle = {The Proceedings of the European Conference on Computer Vision (ECCV)},
  year = {2020}
}

@article{avid,
  title={AVID: Any-Length Video Inpainting with Diffusion Model},
  author={Zhang, Zhixing and Wu, Bichen and Wang, Xiaoyan and Luo, Yaqiao and Zhang, Luxin and Zhao, Yinan and Vajda, Peter and Metaxas, Dimitris and Yu, Licheng},
  journal={arXiv preprint arXiv:2312.03816},
  year={2023}
}

@inproceedings{dynamicshadow,
  title={Dynamic Shadow Unveils Invisible Semantics for Video Outpainting},
  author={Li, Ruilin and Yu, Hang and Qiu, Jiayan},
  booktitle={Advances in Neural Information Processing Systems (NeurIPS)},
  year={2025}
}

@inproceedings{m3ddmplus,
  title={M3DDM+: An improved video outpainting by a modified masking strategy},
  author={Murakawa, Takuya and Fukuzawa, Takumi and Ding, Ning and Tamaki, Toru},
  booktitle={Proceedings of the International Workshop on Advanced Imaging Technology (IWAIT)},
  year={2026}
}


\clearpage
\appendix
{\LARGE\bfseries Supplementary Material\par}
\vspace{1em}

We have attached a \link{\href{https://youtu.be/C-XQCRkbv5E}{\underline{\textbf{video}}}} as Supplementary material, containing video outpainting results and comparisons that include the contents mentioned in the main paper.

\section*{Contents}
\begin{spacing}{1.1}
\link{
\noindent \hyperref[sec:tiling]{A\quad Details of Spatio-Temporal Tiling Based Denoising}\\
\hyperref[sec:gcg]{B\quad Multi-scale \gcg{} Construction}\\
\hspace*{1.5em}\hyperref[subsec:gcg_inference]{B.1\quad Inference for Multi-scale \gcg{}}\\
\hspace*{1.5em}\hyperref[subsec:gcg_training]{B.2\quad Training for Multi-scale \gcg{}}\\
\hyperref[sec:implementation]{C\quad Implementation Details}\\
\hspace*{1.5em}\hyperref[subsec:training_dataset]{C.1\quad Training Dataset}\\
\hspace*{1.5em}\hyperref[subsec:lora_training]{C.2\quad Stage-wise LoRA Training}\\
\hyperref[sec:hyperparameter]{D\quad Hyperparameter Analysis}\\
\hspace*{1.5em}\hyperref[subsec:frame_swapping_schedule]{D.1\quad Global-Local Frame Swapping Schedule}\\
\hspace*{1.5em}\hyperref[subsec:stride_selection]{D.2\quad Stride Selection}\\
\hspace*{1.5em}\hyperref[subsec:keyframe_interval]{D.3\quad Keyframe Interval}\\
\hyperref[sec:ablation]{E\quad Ablation on Spatial and Temporal Compression in \gcg{}}\\
\hspace*{1.5em}\hyperref[subsec:spatial]{E.1\quad Spatially-compressed \gcg{}}\\
\hspace*{1.5em}\hyperref[subsec:temporal]{E.2\quad Temporally-compressed \gcg{}}\\
\hspace*{1.5em}\hyperref[subsec:quantitative]{E.3\quad Quantitative Analysis}\\
\hyperref[sec:autoregressive]{F\quad Discussion on Autoregressive Formulation}\\
\hyperref[sec:detailed_implementation]{G\quad Detailed Implementation}\\
\hspace*{1.5em}\hyperref[subsec:training]{G.1\quad Training}\\
\hspace*{1.5em}\hyperref[subsec:inference]{G.2\quad Inference}\\
\hyperref[sec:scbc_metrics]{H\quad Analysis of SC and BC Metrics}\\
\hyperref[sec:inference_time]{I\quad Inference Time Comparison}\\
\hyperref[sec:user_study]{J\quad User Study}\\
\hyperref[sec:rope_temporal_dimension]{K\quad RoPE Temporal Dimension}\\
\hyperref[sec:dataset]{L\quad Dataset Details}
}
\end{spacing}

\section{Details of Spatio-Temporal Tiling Based Denoising}
\label{sec:tiling}

Applying a diffusion model directly to high-resolution or long videos is computationally expensive. 
To address this, we divide the input video $x$ into a set of overlapping spatio-temporal tiles $\{x^{(i)}\}$, 
where each tile covers a local spatial region and a short temporal window.
Each tile is independently processed by the diffusion model, producing denoised outputs $\{\hat{x}^{(i)}\}$.
To reduce boundary artifacts, tiles are constructed with overlaps in both space and time.
The final video $\hat{x}$ is obtained by blending predictions in the overlapping regions. 
Specifically, for each location $u$ (spatial-temporal coordinate), we compute
\[
\hat{x}(u) = \frac{\sum_{i \in \mathcal{T}(u)} w_i(u)\,\hat{x}^{(i)}(u)}
{\sum_{i \in \mathcal{T}(u)} w_i(u)},
\]
where $\mathcal{T}(u)$ denotes the set of tiles covering $u$, and $w_i(u)$ is a weighting function that assigns higher values near the center of each tile and lower values near the boundaries.
This simple tiling-and-blending strategy allows the model to process videos of arbitrary resolution and length while maintaining smooth transitions across tiles.

\begin{algorithm}[t]
\caption{Multi-scale \gcg{} Construction}
\label{alg:scaffolddensification}

\KwIn{
Initial $\mathcal{G}=\{ \hat{\mathbf{I}}_{k_i}^{\downarrow} \}_{k_i \in \mathcal{K}}$ with keyframe index set $\mathcal{K}=\{k_i\}_{i=1}^{K}$,
Spatially downsampled input video $\mathcal{I}^{\downarrow}=\{\mathbf{I}_f^{\downarrow}\}_{f=1}^{F}$,
Maximum number of frames per forward pass $K$,
Predefined threshold $\tau$,
Video diffusion outpainting model $P$,
}
\KwOut{Refined $\mathcal{I}_{\mathrm{\gcg{}}}$}

\SetKwFunction{FMaxGap}{MaxIndexGap}
\SetKwFunction{FInsert}{RefineFrames}
\SetKwFunction{TS}{TemporalSplit}
\SetKwFunction{TM}{TemporalMerge}
\SetKwFunction{FOutpaint}{VideoOutpainting}

\vspace{0.5em}
\SetKwProg{Fn}{Function}{:}{}
\Fn{\FMaxGap{$\mathcal{K}$}}{
    $\mathcal{K} \leftarrow \operatorname{SortAscending} (\mathcal{K})$ \\
    \Return{$\displaystyle \max\{k_{i+1} - k_i \;|\; i=1,\dots,|\mathcal{K}|-1\}$}
}

\vspace{0.5em}
\Fn{\FInsert{$\mathcal{G}, \mathcal{K}, \mathcal{I}^{\downarrow}$}}{
    \For{$i \gets 1$ \KwTo $|\mathcal{K}|-1$}{
        $k_{\mathrm{mid}} \leftarrow \lfloor (k_i + k_{i+1})/2 \rfloor$ \\
        $\mathcal{K} \gets \mathcal{K} \cup \{ k_{\mathrm{mid}} \}$ \\
        $\mathcal{G} \gets \mathcal{G} \cup \{ \mathbf{I}_{k_{\mathrm{mid}}}^{\downarrow} \}$ \\
    }
    $\mathcal{K} \leftarrow \operatorname{SortAscending} (\mathcal{K})$ \\
    $\mathcal{G} \gets \operatorname{Reorder}(\mathcal{G}, \mathcal{K})$ \\
    \Return{$\mathcal{G}, \mathcal{K}$}
}

\vspace{0.5em}
\While{$\FMaxGap(\mathcal{K}) > \tau$}{
    $\mathcal{G}, \mathcal{K} \leftarrow
    \FInsert(\mathcal{G}, \mathcal{K}, \mathcal{I}^{\downarrow})$ \\
    
    $\mathcal{G} \leftarrow
    \FOutpaint(P, \mathcal{G})$ \\
}
    $\mathcal{I}_{\mathrm{\gcg{}}} = \mathcal{G}$ \\
\Return{$\mathcal{I}_{\mathrm{\gcg{}}}$}
\end{algorithm}

\section{Multi-scale \gcg{} Construction}
\label{sec:gcg}
\subsection{Inference for Multi-scale \gcg.}
\label{subsec:gcg_inference}
For handling significantly long videos, the initial $\mathcal{G}$ constructed from uniformly sampled keyframes may contain excessively large temporal gaps between adjacent keyframes, which limits its effectiveness as a global reference for subsequent dense video outpainting. To resolve this issue, \ours{} iteratively update $\mathcal{G}$ until the maximum temporal distance between adjacent keyframes falls below a predefined threshold. The refinement of the guidance follows the steps in Algorithm~\ref{alg:scaffolddensification}. 
Starting from an initial keyframe index set $\mathcal{K}$, the algorithm repeatedly evaluates the maximum temporal distance between neighboring keyframes and inserts new keyframes at the temporal midpoints of each adjacent keyframe pair whenever this distance exceeds a predefined threshold $\tau$. After each refinement step, the expanded $\mathcal{G}$ is temporally reordered, split into overlapping segments that respect the maximum input length of the video diffusion model, and refined via diffusion-based video outpainting, where previously generated guidance frames are used as fixed references to guide the synthesis of newly inserted frames. This refinement cycle continues until all adjacent keyframe intervals fall below $\tau$, resulting in a temporally dense and globally coherent $\mathcal{I}_{\mathrm{\gcg}}$ that provides stable and consistent guidance for long-range, high-resolution video outpainting.
\fix{\subsection{Training for Multi-scale \gcg.}}
\label{subsec:gcg_training}
\fix{
To enable the model to handle multi-scale temporal sparsity, we simulate sparse keyframe conditions during training. Specifically, given a densely sampled video, we randomly sample a subset of keyframes and remove intermediate frames between them, creating large temporal gaps. This forces the model to learn to construct a coherent guidance even when conditioning frames are sparsely distributed in time. By varying the sampling interval, the model is exposed to multiple levels of temporal sparsity, which improves robustness when constructing the \gcg{} under different temporal scales at inference time.
}

\section{Implementation Details}
\label{sec:implementation}
\subsection{Training Dataset}
\label{subsec:training_dataset}
The model is trained on approximately 17,000 videos sampled from the public OpenVid-1M~\cite{openvid} dataset, with each video resampled to a resolution of 768×768 and 49 frames. Training relies exclusively on the original videos and masked inputs to learn video outpainting. During each iteration, randomly positioned masks with varying scales and shapes are applied from one of the four directions (top, bottom, left, or right), and the model is trained to reconstruct the masked regions in a visually and temporally consistent manner. This masking strategy enables the model to learn background generation and spatial extrapolation under diverse expansion directions and extents.
Since OpenVid-1M predominantly contains static scenes with limited camera or object motion, this limitation is addressed by additionally incorporating 270 videos from the REDS~\cite{reds} (Realistic and Dynamic Scenes) dataset. REDS includes rich camera movements, diverse object dynamics, and high-frame-rate recordings at 120 fps, providing abundant spatio-temporal variations. Joint training on OpenVid-1M and REDS allows the model to better capture complex motion patterns, viewpoint changes, and camera-induced variations, thereby improving robustness and generalization to realistic dynamic video outpainting scenarios.
\subsection{Stage-wise LoRA Training}
\label{subsec:lora_training}
The inference pipeline consists of two stages that share a frozen video diffusion transformer backbone~\cite{wan2025} but use stage-specific LoRA~\cite{lora} modules. In the \first{} stage, a dedicated LoRA optimized for frames that are compressed only along the spatial dimension is applied, whereas in the \second{} stage, a different LoRA optimized for frames compressed jointly along the spatial and temporal dimensions is used. This design allows a single backbone to efficiently adapt to different latent distributions and processing objectives across stages.

The 3D VAE~\cite{wan2025} that used in the model is originally designed to compress inputs by a factor of 8 along the spatial dimensions and approximately 4 along the temporal dimension, under the assumption that adjacent frames are highly correlated. However, the \first{} stage processes sparsely sampled keyframes with limited temporal continuity. Applying the standard spatio-temporal compression to such inputs would force unrelated frames to be merged along the temporal axis, resulting in information dilution and degraded reconstruction quality.
To address this issue without retraining the VAE, the proposed method exploits a structural property of the 3D VAE: temporal compression is not applied when a single frame is provided as input. In the \first{} stage, each frame is therefore encoded independently using spatial-only compression, and the resulting latent representations are concatenated along the temporal axis. This enables precise encoding and decoding of temporally discontinuous frames and provides a representation space well suited for frame-wise background generation.

The \first{} stage is trained to support stable background generation both with and without keyframe conditioning, even when frame intervals vary significantly. In contrast, the \second{} stage operates on temporally dense frame sequences using the standard spatio-temporal VAE and its corresponding LoRA, allowing effective joint modeling of motion and scene dynamics. Keyframes generated in the \first{} stage may serve as temporal anchors in the \second{} stage, and the model ultimately achieves temporally consistent background generation regardless of keyframe availability.

\section{Hyperparameter Analysis}
\label{sec:hyperparameter}
\subsection{Global-Local Frame Swapping Schedule.}
\label{subsec:frame_swapping_schedule}
Global-Local Frame Swapping is designed to correct structural inconsistencies by propagating local temporal cues into global keyframes. This mechanism is most effective during the early denoising stage, where the global structure of the video is established. In later stages, the diffusion process mainly focuses on refining fine details, and applying \swap{} at this stage provides limited benefit.

To analyze this effect, we vary the number of denoising steps during which \swap{} is applied. As shown in Table~\ref{tab:swap_schedule}, applying \swap{} during the first 8 out of 40 denoising steps achieves the best overall performance. Applying it for too few steps fails to sufficiently correct structural inconsistencies, while applying it for too many steps can disrupt fine-detail refinement. This result supports our design choice of limiting \swap{} to early denoising steps.
Moreover, the performance trend shows that \swap{} is more beneficial when used as a coarse structural alignment mechanism rather than a persistent intervention throughout sampling.
This observation suggests that separating early-stage structure correction from late-stage detail synthesis is important for stable video generation.
\begin{table}[t]
\caption{\fix{
Effect of \swap{} schedule. We vary the number of denoising steps where swapping is applied (out of 40 total steps). Best results are shown in \textbf{bold}.
}}
\scalebox{0.90}{
\fix{\begin{tabular}{c|cccccc}
\hline
Swap steps / total 
& PSNR$\uparrow$ 
& SSIM$\uparrow$ 
& FVD$\downarrow$ 
& SC$\uparrow$ 
& BC$\uparrow$ 
& AQ$\uparrow$ \\
\hline\hline
0 / 40  & 16.21 & 0.630 & 141.8  & 0.887 & 0.911 & 0.551 \\
4 / 40  & 15.93 & 0.580 & 196.06 & 0.883 & 0.907 & \textbf{0.612} \\
\rowcolor{yellow!30}
8 / 40  & \textbf{16.60} & 0.633 & \textbf{133.2} & \textbf{0.889} & \textbf{0.912} & 0.555 \\
16 / 40 & 16.21 & \textbf{0.649} & 211.27 & 0.884 & 0.912 & 0.556 \\
\hline
\end{tabular}
}}
\label{tab:swap_schedule}
\end{table}

\subsection{Stride Selection.}
\label{subsec:stride_selection}
The temporal stride used to construct local windows is determined based on the degree of motion in the input video.
In our implementation, this stride is selected via visual inspection; however, it can also be automatically determined by computing optical flow between consecutive frames and using the average magnitude of the flow vectors as a motion score.
For example, a smaller stride can be assigned to videos with a high motion score to capture fast motion more finely, whereas a larger stride can be used for videos with a low motion score to reduce unnecessary temporal redundancy.
We observe that the performance of our method is largely insensitive to the exact choice of stride, as long as it reasonably reflects the motion dynamics of the scene.

\subsection{Keyframe Interval.}
\label{subsec:keyframe_interval}
Keyframe interval controls the temporal spacing between keyframes used in \gcg{} construction. If the interval is too large, temporal gaps increase and important structural inconsistencies may be missed. Conversely, if the interval is too small, redundant keyframes are introduced without improving performance, leading to unnecessary computational overhead. In practice, we find that an interval of 20 provides a good balance between temporal coverage and efficiency.

\section{Ablation on Spatial and Temporal Compression in \gcg{}}
\label{sec:ablation}
\subsection{Spatially-compressed \gcg{}}
\label{subsec:spatial}
To analyze the impact of temporal compression, we implement the Spatially-compressed \gcg{} by disabling the keyframe-first processing and applying only spatial bicubic downsampling to all frames. Tile-based diffusion sampling is directly performed over overlapping temporal windows at the reduced resolution, followed by bicubic upsampling and the same spatial refinement as in our \second{} stage. 
While spatial structures are generally plausible, this setting suffers from long-term temporal inconsistency due to the absence of a global temporal anchor across windows.
\subsection{Temporally-compressed \gcg{}}
\label{subsec:temporal}
To analyze the impact of spatial compression, we implement the Temporally-compressed \gcg{} by removing spatial downsampling while retaining keyframe sampling. Keyframe outpainting is performed using overlapping spatial tiling, and the generated keyframes are then inserted as temporal anchors for spatio-temporal tiled completion of the full sequence. 
Without a globally compressed spatial guidance, this setting often exhibits repetition artifacts and structural inconsistency across adjacent regions, even though temporal anchors help preserve coarse long-range temporal structure.
\fix{\subsection{Quantitative Analysis}
\label{subsec:quantitative}
To quantitatively evaluate the contribution of spatial and temporal compression in the \gcg, we compare spatial-only compression, temporal-only compression, and full spatio-temporal compression. As shown in Table~\ref{tab:ablation_compression}, combining both spatial and temporal compression yields the best overall performance across most metrics.
Spatial-only compression improves visual fidelity (PSNR, SSIM), whereas temporal-only compression enhances temporal consistency (SC). Our full model achieves the best balance between these factors, resulting in the strongest overall performance.}

\begin{table}[H]
\caption{\fix{
Quantitative ablation on spatial and temporal compression in \gcg. Best results are shown in \textbf{bold}.
}}
\scalebox{0.90}{
\fix{\begin{tabular}{c|cccccc}
\hline
Method & PSNR$\uparrow$ & SSIM$\uparrow$ & FVD$\downarrow$ & SC$\uparrow$ & BC$\uparrow$ & AQ$\uparrow$ \\\hline\hline
Baseline & 12.67 & 0.510 & 1860.2 & 0.837 & 0.872 & 0.435 \\
Spatial-only & 15.08 & 0.600 & 593.3 & 0.870 & 0.901 & 0.516 \\
Temporal-only & 12.77 & 0.531 & 1361 & \textbf{0.889} & 0.898 & 0.519 \\
\rowcolor{yellow!30}
Ours & \textbf{15.32} & \textbf{0.620} & \textbf{564.6} & 0.877 & \textbf{0.901} & \textbf{0.520} \\\hline
\end{tabular}
}}
\label{tab:ablation_compression}
\end{table}

\section{Discussion on Autoregressive Formulation.}
\label{sec:autoregressive}
One may consider building our framework on top of an autoregressive video generation model~\cite{ar,videogpt,phenaki,gao2024ca2,yin2025slow}. Autoregressive models have demonstrated strong temporal modeling capabilities and can generate long video sequences by progressively conditioning each frame on previously generated frames. This sequential formulation allows the model to accumulate temporal context over time and to capture plausible motion and appearance transitions. However, it inherently assumes a causal temporal order: each frame is generated using only past observations, while future frames remain unavailable at the time of generation. Although this assumption is well suited for video prediction or forward video synthesis, it is fundamentally misaligned with video outpainting, where the missing regions should be inferred from visual evidence distributed across the entire sequence.

In particular, video outpainting frequently involves non-causal dependencies. Due to camera motion, object motion, or changes in visibility, scene content that is missing or outside the image boundary in an earlier frame may become visible in later frames. In such cases, the outpainted regions in earlier frames should be consistent with the visual evidence observed in future frames. Autoregressive generation cannot directly exploit such future information, and conditioning only on past frames may therefore produce outpainted content that conflicts with later observations in terms of scene layout, texture, object geometry, or appearance. These conflicts can result in severe temporal inconsistencies and flickering artifacts across the generated video. Our method avoids this limitation by constructing a \gcg{} that captures global spatio-temporal structure over the entire sequence, and by generating frames jointly rather than sequentially. This enables our framework to leverage both past and future visual evidence, leading to outpainted regions that remain more coherent and temporally consistent throughout the video.

\section{Detailed Implementation}
\label{sec:detailed_implementation}

We provide detailed pseudocode for both the inference and training procedures to improve the clarity and reproducibility of our method.

\begin{algorithm}[t]
\caption{\fix{HL-OutPaint Training}}
\color{black}
\label{alg:hloutpaint_training}

\KwIn{
Training dataset $\mathcal{D}$,
training stage $s \in \{1,2\}$,
pretrained diffusion model $P$
}
\KwOut{Trained LoRA parameters}

\SetKwFunction{FInsert}{InsertLoRA}
\SetKwFunction{FStageLoss}{StageWiseSamplePreparationAndLoss}

\FInsert the LoRA modules into the DiT blocks of $P$ \;

\ForEach{$(\mathcal{I}, \mathbf{p}) \sim \mathcal{D}$}{
    $\mathcal{L} \leftarrow \FStageLoss(\mathcal{I}, \mathbf{p}, s, P)$ \;
    Update the LoRA parameters using $\nabla \mathcal{L}$ \;
}

\Return{trained LoRA parameters}
\end{algorithm}

\subsection{Training}
\label{subsec:training}

For training, we adopt a two-stage optimization strategy for HL-OutPaint.
As shown in \cref{alg:hloutpaint_training}, we first insert LoRA modules into the DiT blocks of the pretrained diffusion model and optimize only these newly introduced parameters.
For each training sample, the algorithm delegates the stage-dependent sample preparation and diffusion loss computation to the stage-wise procedure in \cref{alg:hloutpaint_stage_and_loss}.
This design allows the pretrained video diffusion prior to be efficiently adapted to hierarchical long-video outpainting while keeping the majority of the original model parameters frozen.

\subsubsection{Stage-wise sample construction and objective}
\label{subsubsec:training_stage_objective}

The detailed sample construction and objective are described in \cref{alg:hloutpaint_stage_and_loss}.
Given an input video and its text prompt, the procedure prepares the target video, masked conditioning video, binary mask, and anchor frames according to the current training stage.
It then encodes the target and conditioning inputs into latent space, applies temporal cropping, injects diffusion noise at a sampled timestep, and computes the velocity-prediction loss with scheduler-dependent weighting.
The resulting loss is returned to the training loop in \cref{alg:hloutpaint_training}, where it is used to update the LoRA parameters.
\begin{algorithm}[t]
\caption{\fix{Stage-wise Sample Preparation and Loss}}
\color{black}
\label{alg:hloutpaint_stage_and_loss}

\KwIn{
Original video $\mathcal{I}=\{\mathbf{I}_f\}_{f=1}^{F}$,
prompt $\mathbf{p}$,
training stage $s$,
diffusion model $P$
}
\KwOut{Training loss $\mathcal{L}$}

\SetKwFunction{FMask}{MakeTrainingMask}
\SetKwFunction{FAnchor}{GenerateAnchorFrames}
\SetKwFunction{FEncodeT}{EncodeTarget}
\SetKwFunction{FEncodeC}{EncodeCondition}
\SetKwFunction{FCrop}{TemporalCrop}
\SetKwFunction{FNoisy}{AddNoise}
\SetKwFunction{FTarget}{SchedulerTarget}
\SetKwFunction{FWeight}{SchedulerWeight}

\eIf{$s = 1$}{
    Select $13$ evenly spaced keyframes from $\mathcal{I}$ and use them as $\mathcal{I}_{\mathrm{target}}$ \;
    $(\bar{\mathcal{I}}, \mathcal{M}) \leftarrow \FMask(\mathcal{I}_{\mathrm{target}})$ \;
    $\mathcal{K} \leftarrow \FAnchor(\mathcal{I}_{\mathrm{target}})$ with short temporal stride \;
    Replace the masked anchor frames with the ground-truth frames and set the corresponding masks to one \;
    Encode the target and masked videos with independent frame-wise VAE processing \;
}{
    $\mathcal{I}_{\mathrm{target}} \leftarrow \mathcal{I}$ \;
    $(\bar{\mathcal{I}}, \mathcal{M}) \leftarrow \FMask(\mathcal{I}_{\mathrm{target}})$ \;
    $\mathcal{K} \leftarrow \FAnchor(\mathcal{I}_{\mathrm{target}})$ with longer temporal stride \;
    Replace the masked anchor frames with the ground-truth frames and set the corresponding masks to one \;
    Encode the target and masked videos with the standard temporally-compressed video VAE \;
}

$\mathbf{z}_{\mathrm{target}} \leftarrow \FEncodeT(\mathcal{I}_{\mathrm{target}})$ \;
$\mathbf{y} \leftarrow \FEncodeC(\bar{\mathcal{I}}, \mathcal{M})$ \;
Concatenate the downsampled mask channels and the masked-video latent channels to form the condition tensor \;

$\mathbf{z}_{\mathrm{target}}, \mathbf{y} \leftarrow \FCrop(\mathbf{z}_{\mathrm{target}}, \mathbf{y})$ \;
Sample timestep $t$ and Gaussian noise $\boldsymbol{\epsilon}$ \;
$\mathbf{z}_{t} \leftarrow \FNoisy(\mathbf{z}_{\mathrm{target}}, \boldsymbol{\epsilon}, t)$ \;
$\mathbf{v}^{\star} \leftarrow \FTarget(\mathbf{z}_{\mathrm{target}}, \boldsymbol{\epsilon}, t)$ \;
$\hat{\mathbf{v}} \leftarrow P(\mathbf{z}_{t}, \mathbf{y}, \mathbf{p}, t)$ \;
$\mathcal{L} \leftarrow \|\hat{\mathbf{v}} - \mathbf{v}^{\star}\|_2^2 \cdot \FWeight(t)$ \;

\Return{$\mathcal{L}$}
\end{algorithm}

\subsubsection{Two-stage training strategy}
\label{subsubsec:training_two_stage}

In the first stage of \cref{alg:hloutpaint_stage_and_loss}, we select $13$ evenly spaced keyframes from the original video as the target sequence.
This stage focuses on establishing reliable spatial completion behavior from sparsely sampled keyframes.
The target and masked videos are encoded with independent frame-wise VAE processing, which encourages the LoRA-adapted DiT blocks to learn high-quality boundary extrapolation and mask-conditioned reconstruction without being overly constrained by long-range temporal compression.
Short-stride anchor frames are further used as sparse temporal references, and the masked anchor positions are replaced with their corresponding ground-truth frames to stabilize reconstruction around reliable observations.

In the second stage of \cref{alg:hloutpaint_stage_and_loss}, we train on the full video clip using the standard temporally-compressed video VAE.
Compared with the first stage, this stage uses anchor frames with a longer temporal stride, enabling the model to extend the spatial outpainting capability learned from keyframes to longer video sequences.
This stage encourages global motion coherence while preserving consistency with the visible regions and the provided textual prompt.

\subsubsection{Conditioning and optimization}
\label{subsubsec:training_conditioning_optimization}

Across both stages, the masked-video latents and downsampled mask channels are concatenated as explicit conditioning inputs, as specified in \cref{alg:hloutpaint_stage_and_loss}.
This conditioning allows the diffusion model to distinguish observed, masked, and anchor-provided regions throughout the denoising process.
The anchor-frame replacement strategy provides sparse but reliable temporal references, reducing drift across long sequences and stabilizing training when the outpainted regions span large spatial extents.

Finally, the diffusion objective in \cref{alg:hloutpaint_stage_and_loss} follows the velocity-prediction formulation of the underlying scheduler.
After noise is added to the target latent, the model predicts the scheduler target conditioned on the masked video, mask, prompt, and timestep.
The weighted squared error loss is then used in \cref{alg:hloutpaint_training} to update only the inserted LoRA parameters, enabling efficient adaptation of the pretrained video diffusion model to hierarchical long-video outpainting.

\begin{algorithm}[t]
\caption{\fix{HL-OutPaint Inference}}
\color{black}
\label{alg:hloutpaint_inference}

\KwIn{
Input video $\mathcal{I}=\{\mathbf{I}_f\}_{f=1}^{F}$,
target resolution $(H,W)$,
initial keyframe count $K$,
threshold $\tau$,
stage-1 LoRA $(L_1^{h}, L_1^{l})$,
stage-2 LoRA $(L_2^{h}, L_2^{l})$,
video diffusion outpainting pipeline $P$
}
\KwOut{Final outpainted video $\hat{\mathcal{I}}$}

\SetKwFunction{FPad}{PadVideoLength}
\SetKwFunction{FMask}{MakeInferenceMask}
\SetKwFunction{FGuide}{ResizeForGuidance}
\SetKwFunction{FKey}{EvenKeyframeIndices}
\SetKwFunction{FMaxGap}{MaxIndexGap}
\SetKwFunction{FMid}{MidIndices}
\SetKwFunction{FSparse}{SparseGuidance}
\SetKwFunction{FInsert}{InsertGeneratedFrames}
\SetKwFunction{FDense}{DenseGuidance}
\SetKwFunction{FRestore}{RestoreGuidanceResolution}
\SetKwFunction{FFinal}{FinalOutpainting}

\vspace{0.5em}
\SetKwProg{Fn}{Function}{:}{}
\Fn{\FMaxGap{$\mathcal{K}$}}{
    $\mathcal{K} \leftarrow \operatorname{SortAscending}(\mathcal{K})$ \;
    \Return{$\displaystyle \max\{k_{i+1}-k_i \;|\; i=1,\dots,|\mathcal{K}|-1\}$}
}

\vspace{0.5em}
\Fn{\FMid{$\mathcal{K}, \tau$}}{
    $\mathcal{S} \leftarrow \emptyset$ \;
    \For{$i \gets 1$ \KwTo $|\mathcal{K}|-1$}{
        \If{$k_{i+1}-k_i > \tau$}{
            $k_{\mathrm{mid}} \leftarrow \lfloor (k_i + k_{i+1})/2 \rfloor$ \;
            $\mathcal{S} \leftarrow \mathcal{S} \cup \{k_{\mathrm{mid}}\}$ \;
        }
    }
    \Return{$\operatorname{SortAscending}(\mathcal{S})$}
}

\vspace{0.5em}
\Fn{\FInsert{$\bar{\mathcal{I}}^{\downarrow}, \mathcal{M}^{\downarrow}, \mathcal{G}, \mathcal{K}$}}{
    \ForEach{$k \in \mathcal{K}$}{
        Replace the $k$-th frame of $\bar{\mathcal{I}}^{\downarrow}$ with the generated guidance frame in $\mathcal{G}$ \;
        Set the $k$-th mask in $\mathcal{M}^{\downarrow}$ to one \;
    }
    \Return{$\bar{\mathcal{I}}^{\downarrow}, \mathcal{M}^{\downarrow}$}
}

\vspace{0.5em}
$(\mathcal{I}, F_{\mathrm{orig}}) \leftarrow \FPad(\mathcal{I})$ \;
$(\bar{\mathcal{I}}, \mathcal{M}) \leftarrow \FMask(\mathcal{I}, H, W)$ \;
$(\bar{\mathcal{I}}^{\downarrow}, \mathcal{M}^{\downarrow}) \leftarrow \FGuide(\bar{\mathcal{I}}, \mathcal{M})$ \;
$\mathcal{K} \leftarrow \FKey(F, K)$ \;

$\mathcal{G} \leftarrow \FSparse(P, \bar{\mathcal{I}}^{\downarrow}, \mathcal{M}^{\downarrow}, \mathcal{K}, L_1^{h}, L_1^{l})$ \;

\While{$\FMaxGap(\mathcal{K}) > \tau$}{
    $\mathcal{S} \leftarrow \FMid(\mathcal{K}, \tau)$ \;
    $\bar{\mathcal{I}}^{\downarrow}, \mathcal{M}^{\downarrow} \leftarrow \FInsert(\bar{\mathcal{I}}^{\downarrow}, \mathcal{M}^{\downarrow}, \mathcal{G}, \mathcal{K})$ \;
    Re-run sparse guidance on the active keyframe set $\mathcal{K} \cup \mathcal{S}$ using the stage-1 LoRA \;
    Update $\mathcal{G}$ with the newly synthesized midpoint frames \;
    $\mathcal{K} \leftarrow \operatorname{SortAscending}(\mathcal{K} \cup \mathcal{S})$ \;
}

$\bar{\mathcal{I}}^{\downarrow}, \mathcal{M}^{\downarrow} \leftarrow \FInsert(\bar{\mathcal{I}}^{\downarrow}, \mathcal{M}^{\downarrow}, \mathcal{G}, \mathcal{K})$ \;
$\tilde{\mathcal{I}}_{\mathrm{GCS}} \leftarrow \FDense(P, \bar{\mathcal{I}}^{\downarrow}, \mathcal{M}^{\downarrow}, \mathcal{G}, L_2^{h}, L_2^{l})$ \;
$\mathcal{I}_{\mathrm{GCS}} \leftarrow \FRestore(\tilde{\mathcal{I}}_{\mathrm{GCS}}, H, W)$ \;
$\hat{\mathcal{I}} \leftarrow \FFinal(P, \bar{\mathcal{I}}, \mathcal{M}, \mathcal{I}_{\mathrm{GCS}}, L_2^{h}, L_2^{l})$ \;
Trim the padded tail frames so that the output length matches $F_{\mathrm{orig}}$ \;

\Return{$\hat{\mathcal{I}}$}
\end{algorithm}
\begin{algorithm}[t]
\caption{\fix{Core Pipeline Forward}}
\color{black}
\label{alg:hloutpaint_core_forward}

\KwIn{
Masked video $\bar{\mathcal{I}}$,
mask $\mathcal{M}$,
optional dense guidance $\mathcal{I}_{\mathrm{GCS}}$,
optional keyframe set $\mathcal{K}$,
diffusion pipeline $P$
}
\KwOut{Generated video}

\SetKwFunction{FPrompt}{EncodePrompt}
\SetKwFunction{FEncode}{EncodeCondition}
\SetKwFunction{FInit}{InitializeNoise}
\SetKwFunction{FTarget}{EncodeGuidance}
\SetKwFunction{FWarm}{WarmStart}
\SetKwFunction{FGLA}{GLADiTUpdate}
\SetKwFunction{FPatch}{PatchwiseUpdate}
\SetKwFunction{FStep}{SchedulerStep}
\SetKwFunction{FDecode}{DecodeLatent}

$\mathbf{c} \leftarrow \FPrompt()$ \;
$\mathbf{y} \leftarrow \FEncode(\bar{\mathcal{I}}, \mathcal{M})$ \;
$\mathbf{z}_{T} \leftarrow \FInit()$ \;
Determine the temporal and spatial patch sizes in latent space from the target video size and the VAE downsampling factor \;

\If{$\mathcal{I}_{\mathrm{GCS}}$ is given}{
    $\mathbf{z}_{\mathrm{guide}} \leftarrow \FTarget(\mathcal{I}_{\mathrm{GCS}})$ \;
    Add scheduler-consistent noise to $\mathbf{z}_{\mathrm{guide}}$ and use it as a warm-start prior \;
    $\mathbf{z}_{T} \leftarrow \FWarm(\mathbf{z}_{T}, \mathbf{z}_{\mathrm{guide}})$ \;
}

\For{$t \gets T$ \KwTo $1$}{
    \eIf{$\mathcal{K} \neq \emptyset$}{
        Build one local temporal window around each keyframe in $\mathcal{K}$ \;
        Extract a global keyframe stack using the same keyframe indices \;
        Denoise the local windows with the corresponding local condition slices \;
        Denoise the global stack with the corresponding global condition slices \;
        Merge the corresponding local and global keyframe latents according to the predefined schedule \;
        $\hat{\mathbf{v}}_t \leftarrow \FGLA(P, \mathbf{z}_{t}, \mathbf{y}, \mathbf{c}, \mathcal{K}, t)$ \;
    }{
        Split $\mathbf{z}_{t}$ and $\mathbf{y}$ into temporal windows \;
        Split each temporal window into spatial patches \;
        Denoise every spatio-temporal patch independently with the shared prompt condition $\mathbf{c}$ \;
        Merge the spatial patches and then merge the temporal windows back to the full latent tensor \;
        $\hat{\mathbf{v}}_t \leftarrow \FPatch(P, \mathbf{z}_{t}, \mathbf{y}, \mathbf{c}, t)$ \;
    }
    $\mathbf{z}_{t-1} \leftarrow \FStep(\mathbf{z}_{t}, \hat{\mathbf{v}}_t, t)$ \;
}

\Return{$\FDecode(\mathbf{z}_{0})$}
\end{algorithm}

\subsection{Inference}
\label{subsec:inference}

For inference, \cref{alg:hloutpaint_inference} describes the overall hierarchical long-video outpainting procedure, while \cref{alg:hloutpaint_core_forward} details the core diffusion forward process used inside the sparse and dense generation stages.
Together, these two algorithms define how HL-OutPaint first constructs temporally coherent guidance and then uses it as a spatio-temporal prior for full-resolution video outpainting.

\subsubsection{Hierarchical inference pipeline}
\label{subsubsec:inference_hierarchical_pipeline}

As shown in \cref{alg:hloutpaint_inference}, the inference procedure first pads the input sequence to a valid video length and constructs the masked outpainting condition at the target spatial resolution.
To make long-range generation tractable, the method then builds a low-resolution guidance space and selects an initial set of evenly distributed keyframes.
Starting from these sparse keyframes, stage-1 LoRA is used to synthesize coarse outpainted guidance frames, which serve as globally consistent anchors for the extended spatial regions.

The sparse guidance is progressively densified through an iterative midpoint insertion strategy.
At each iteration, \cref{alg:hloutpaint_inference} identifies temporal intervals whose keyframe gaps exceed the threshold $\tau$, generates additional midpoint guidance frames, and inserts the newly synthesized frames back into the guidance video and mask.
This sparse-to-dense procedure reduces large temporal gaps and allows the generated guidance to propagate smoothly across the full video, rather than relying on a small number of distant anchor frames.
After the keyframe spacing satisfies the threshold, the dense guidance is refined using the stage-2 LoRA and restored to the target resolution.
The restored dense guidance is then passed to the final outpainting stage, where the full-resolution video is generated and the padded tail frames are removed to recover the original video length.

\subsubsection{Core diffusion forward process}
\label{subsubsec:inference_core_forward}

The internal generation process used by the inference pipeline is described in \cref{alg:hloutpaint_core_forward}.
Given a masked video, a binary mask, optional dense guidance, and an optional keyframe set, the core pipeline first encodes the text prompt and the masked video condition, and initializes the diffusion latent with noise.
When dense guidance $\mathcal{I}_{\mathrm{GCS}}$ is available, the pipeline encodes it into the latent space, injects scheduler-consistent noise, and uses the resulting latent as a warm-start prior.
This allows the final generation to follow the globally consistent guidance produced by \cref{alg:hloutpaint_inference} while still allowing the diffusion model to refine local details.

During denoising, \cref{alg:hloutpaint_core_forward} switches between two update modes depending on whether a keyframe set is provided.
When keyframes are available, the pipeline performs keyframe-aware global-local latent aggregation: it builds local temporal windows around the selected keyframes, extracts a global keyframe stack, denoises both views, and merges the corresponding latent predictions according to the aggregation schedule.
This mode is used for guided sparse generation, where global consistency across distant frames is critical.
When no keyframe set is provided, the pipeline instead performs patchwise spatio-temporal denoising by splitting the latent video into temporal windows and spatial patches, denoising each patch independently, and merging them back into the full latent tensor.
This mode enables dense full-video synthesis at high resolution while keeping memory consumption manageable.

Overall, \cref{alg:hloutpaint_inference} defines the hierarchical sparse-to-dense guidance construction, whereas \cref{alg:hloutpaint_core_forward} defines the reusable diffusion forward process that performs either keyframe-aware global-local aggregation or dense patchwise denoising.
This separation allows HL-OutPaint to preserve long-range temporal structure through guidance while maintaining local spatial detail in the final outpainted regions.

\fix{\section{Analysis of SC and BC Metrics}
\label{sec:scbc_metrics}
SC and BC are commonly used metrics for evaluating temporal consistency by measuring similarity in global feature representations across frames. However, because they rely on one-dimensional global features, they are relatively insensitive to spatial structural differences within each frame. To better capture spatially localized consistency, we adopt a spatially-aware evaluation strategy. Specifically, each frame is divided into a set of spatial tiles, and SC and BC are computed independently for each tile and then averaged. This allows the metrics to reflect fine-grained spatial variations that are otherwise smoothed out in global representations. When comparing our method with VACE~\cite{vace}, we observe that the performance gap appears small under the standard global evaluation. However, under the tiled evaluation, the gap becomes significantly larger. In particular, the difference in SC increases from 0.0009 to 0.0132 (approximately 14.7$\times$), and the difference in BC increases from 0.0004 to 0.0041 (approximately 10$\times$). These results indicate that our method achieves stronger spatially consistent generation compared to VACE, which is not fully captured by global feature-based metrics.}

\section{Inference Time Comparison}
\label{sec:inference_time}
We compare the inference time of different video outpainting methods on a 500-frame $720\times1280$ video using an A100-80GB GPU. As shown in Table~\ref{tab:inference_time}, our method achieves the fastest inference time among all compared methods. In particular, our method requires 105 minutes, outperforming VACE, the second-fastest baseline, which requires 143 minutes. This result demonstrates that the proposed hierarchical guidance construction and efficient stage-wise processing improve not only temporal consistency and visual quality but also inference efficiency for long high-resolution video outpainting.

\begin{table}[t]
\caption{\fix{
Inference time comparison on a 500-frame $720\times1280$ video using an A100-80GB GPU. Best results are shown in \textbf{bold}.
}}
\scalebox{0.90}{
\fix{\begin{tabular}{c|ccccc}
\hline
Method & M3DDM & MOTIA & Infinite-Canvas & VACE & Ours \\\hline\hline
Time (min)$\downarrow$ & 780 & 161 & 285 & 143 & \textbf{105} \\\hline
\end{tabular}}
}
\label{tab:inference_time}
\end{table}

\fix{\section{User Study}
\label{sec:user_study}
We conduct a user study to evaluate perceptual quality of video outpainting results. We recruit 20 participants and evaluate on 10 randomly selected videos. For each video, participants are shown results from M3DDM~\cite{m3ddm}, MOTIA~\cite{motia}, Infinite-Canvas~\cite{ic}, VACE~\cite{vace}, and our method and asked to select the best result for each of the following criteria: visual quality, temporal consistency, subject quality, and background quality. We report the vote percentage for each method in Table~\ref{tab:user_study}. As shown in the table, our method is consistently preferred across all criteria, demonstrating clear advantages in both visual fidelity and temporal coherence.}
\begin{table}[h]
\centering
\caption{\fix{User study results. We report vote percentages across 20 participants and 10 videos.}}
\scalebox{0.90}{
\fix{\begin{tabular}{c|cccc}
\hline
Method & Visual & Temporal & Subject & Background \\
\hline
M3DDM & 0.00 & 0.00 & 0.00 & 0.00 \\
MOTIA & 0.00 & 0.00 & 0.01 & 0.00 \\
Infinite-Canvas & 0.00 & 0.00 & 0.02 & 0.00 \\
VACE & 0.05 & 0.03 & 0.05 & 0.05 \\
\rowcolor{yellow!30}
HL-OutPaint (Ours) & \textbf{0.95} & \textbf{0.97} & \textbf{0.93} & \textbf{0.95} \\
\hline
\end{tabular}}
}
\label{tab:user_study}
\end{table}

\section{RoPE Temporal Dimension.}
\label{sec:rope_temporal_dimension}
Since the first stage operates on temporally sparse keyframes, one may wonder whether special handling is required for the temporal dimension of RoPE. In our method, we do not apply any explicit modification or additional processing to temporal RoPE. From the perspective of the diffusion backbone, the keyframes are processed as a regular frame sequence using the original positional encoding. During fine-tuning, the model sufficiently adapts to the domain difference introduced by sparse keyframe inputs, without requiring manual adjustment of the temporal dimension of RoPE.

\section{Dataset Details}
\label{sec:dataset}

All videos are collected from Pexels~\cite{pexels} (\url{https://www.pexels.com}) under its free license. We provide the corresponding URLs. All data are publicly accessible and are expected to remain available.
The video lists for the long‑video and short‑form datasets are provided in \cref{tab:long_video_dataset,tab:short_form_dataset}.

\begin{table*}[t]
\centering
\small
\fix{\begin{tabular}{c p{11cm}}
\toprule
ID & Video URL \\
\midrule
001 & \url{https://www.pexels.com/video/scenic-bridge-at-sunset-over-tranquil-river-28683111/} \\
002 & \url{https://www.pexels.com/video/snow-covered-mountain-landscape-in-winter-28963324/} \\
003 & \url{https://www.pexels.com/video/driving-through-sunny-urban-overpass-31268032/} \\
004 & \url{https://www.pexels.com/video/historic-mosque-exterior-with-lush-trees-34815636/} \\
005 & \url{https://www.pexels.com/video/bustling-waterfront-promenade-in-vibrant-city-35214543/} \\
006 & \url{https://www.pexels.com/video/bustling-waterfront-walkway-with-boats-35215643/} \\
007 & \url{https://www.pexels.com/video/historical-greek-and-roman-ruins-17675370/} \\
008 & \url{https://www.pexels.com/video/railway-between-buildings-2005977/} \\
009 & \url{https://www.pexels.com/video/double-decker-bus-in-the-city-2235731/} \\
010 & \url{https://www.pexels.com/video/a-homeless-man-hugging-a-dog-8077538/} \\
011 & \url{https://www.pexels.com/video/group-of-people-picking-up-trash-in-a-park-3209571/} \\
012 & \url{https://www.pexels.com/video/a-man-typing-on-his-laptop-7685206/} \\
013 & \url{https://www.pexels.com/video/video-of-city-traffic-at-night-5057439/} \\
014 & \url{https://www.pexels.com/video/low-angle-shot-of-a-man-talking-on-cellphone-5321393/} \\
015 & \url{https://www.pexels.com/video/dried-leaves-in-the-park-with-a-statue-5912265/} \\
016 & \url{https://www.pexels.com/video/close-up-shot-of-bushes-5978808/} \\
017 & \url{https://www.pexels.com/video/call-center-agent-7682895/} \\
018 & \url{https://www.pexels.com/video/man-and-woman-working-together-6876447/} \\
019 & \url{https://www.pexels.com/video/city-road-person-street-7252611/} \\
020 & \url{https://www.pexels.com/video/railway-between-buildings-2005977/} \\
\bottomrule
\end{tabular}}
\caption{Long-Video dataset used in our experiments.}
\label{tab:long_video_dataset}
\end{table*}

\begin{table*}[t]
\centering
\small
\fix{\begin{tabular}{c p{11cm}}
\toprule
ID & Video URL \\
\midrule
000 & \url{https://www.pexels.com/ko-kr/video/35412061/} \\
001 & \url{https://www.pexels.com/ko-kr/video/35609413/} \\
002 & \url{https://www.pexels.com/ko-kr/video/35609378/} \\
003 & \url{https://www.pexels.com/ko-kr/video/35595327/} \\
004 & \url{https://www.pexels.com/ko-kr/video/35608930/} \\
005 & \url{https://www.pexels.com/ko-kr/video/35607780/} \\
006 & \url{https://www.pexels.com/ko-kr/video/35605612/} \\
007 & \url{https://www.pexels.com/ko-kr/video/35350329/} \\
008 & \url{https://www.pexels.com/ko-kr/video/35601783/} \\
\bottomrule
\end{tabular}}
\caption{Short-Form dataset used in our experiments.}
\label{tab:short_form_dataset}
\end{table*}

\clearpage


\end{document}